\documentclass[10pt,twocolumn,letterpaper]{article}

\usepackage{wacv}
\usepackage{times}
\usepackage{epsfig}
\usepackage{graphicx}
\usepackage{amsmath}
\usepackage{amssymb}
\usepackage{booktabs}

\usepackage{comment}
\usepackage{cite}
\usepackage{amsmath,amssymb} 
\usepackage{color}
\usepackage{kotex}
\usepackage{varwidth}
\usepackage{xcolor}

\usepackage{verbatim}
\usepackage{multirow}
\usepackage{booktabs}
\usepackage{adjustbox}
\usepackage{array}
\usepackage{bbm}
\usepackage{mathtools}
\usepackage{grffile}
\usepackage{enumitem}

\newcommand{\PreserveBackslash}[1]{\let\temp=\\#1\let\\=\temp}
\newcolumntype{x}[1]{>{\centering\let\newline\\\arraybackslash\hspace{0pt}}p{#1}}
\newcolumntype{R}[1]{>{\PreserveBackslash\raggedleft}p{#1}}

\def\wacvPaperID{962} 

\wacvalgorithmstrack   

\wacvfinalcopy 

\ifwacvfinal
\usepackage[breaklinks=true,bookmarks=false]{hyperref}
\else
\usepackage[pagebackref=true,breaklinks=true,colorlinks,bookmarks=false]{hyperref}
\fi
\usepackage[symbol]{footmisc}

\usepackage[capitalize]{cleveref}
\crefname{section}{Sec.}{Secs.}
\Crefname{section}{Section}{Sections}
\Crefname{table}{Table}{Tables}
\crefname{table}{Tab.}{Tabs.}

\begin{document}

\definecolor{neon}{RGB}{32, 179, 59}
\definecolor{blackpink}{RGB}{235, 52, 192}
\definecolor{hwanta}{RGB}{235, 155, 52}
\newcommand{\jy}[1]{{\color{neon}#1}}
\newcommand{\mh}[1]{{\color{blackpink}#1}}
\newcommand{\sh}[1]{{\color{hwanta}#1}}
\newcommand{\win}[1]{{\small{\color{blue}#1}}}
\newcommand{\lose}[1]{{\small{\color{red}#1}}}

\def\etc{\emph{etc}\onedot} \def\vs{\emph{vs}\onedot}
\def\smalleq{\texttt{=}}

\title{iColoriT: Towards Propagating Local Hint to the Right Region in Interactive Colorization by Leveraging Vision Transformer}

\author{
Jooyeol Yun*, \quad Sanghyeon Lee*, \quad Minho Park*, \quad Jaegul Choo\\
Korea Advanced Institute of Science and Technology (KAIST)\\
Daejeon, Korea\\
{\tt\small  \{blizzard072, shlee6825, m.park, jchoo\}@kaist.ac.kr}
}

\maketitle
\thispagestyle{empty}

\begin{abstract}
Point-interactive image colorization aims to colorize grayscale images when a user provides the colors for specific locations. 
It is essential for point-interactive colorization methods to appropriately propagate user-provided colors (\ie, user hints) in the entire image to obtain a reasonably colorized image with minimal user effort. 
However, existing approaches often produce partially colorized results due to the inefficient design of stacking convolutional layers to propagate hints to distant relevant regions. 
To address this problem, we present iColoriT, a novel point-interactive colorization Vision Transformer capable of propagating user hints to relevant regions, leveraging the global receptive field of Transformers. 
The self-attention mechanism of Transformers enables iColoriT to selectively colorize relevant regions with only a few local hints. 
Our approach colorizes images in real-time by utilizing pixel shuffling, an efficient upsampling technique that replaces the decoder architecture. 
Also, in order to mitigate the artifacts caused by pixel shuffling with large upsampling ratios, we present the local stabilizing layer. 
Extensive quantitative and qualitative results demonstrate that our approach highly outperforms existing methods for point-interactive colorization, producing accurately colorized images with a user's minimal effort. 
Official codes are available at {\url{https://pmh9960.github.io/research/iColoriT/}.}

\footnotetext[1]{\vspace{-0.7cm} indicates equal contribution.}
\end{abstract}


\vspace{-0.5cm}
\section{Introduction}
\vspace{-0.1cm}

Unconditional image colorization~\cite{coltran, instanceaware, cic, chromagan, Iizuka2016letcolor, memo} has shown remarkable achievement in restoring the vibrance of grayscale photographs or films in a fully-automatic manner. 
Interactive colorization methods~\cite{he2018deep, xu2020stylization,  zhang2019deep, zhang2017, levin2004, side} further extend the task to allow users to generate colorized images with specific color conditions. 
These approaches can dramatically reduce the user effort for producing specific colorized images. 
It can also serve as an effective way of editing photos by re-coloring existing images to have a new color theme. 
Among different types of interactions provided by users (\eg, a reference image or a color palette), point- or scribble-based interactions~\cite{zhang2017, levin2004, side} are designed to progressively colorize images when a user provides the colors at specific point locations.  

\begin{figure}[t]
    \centering
    \includegraphics[width=\columnwidth]{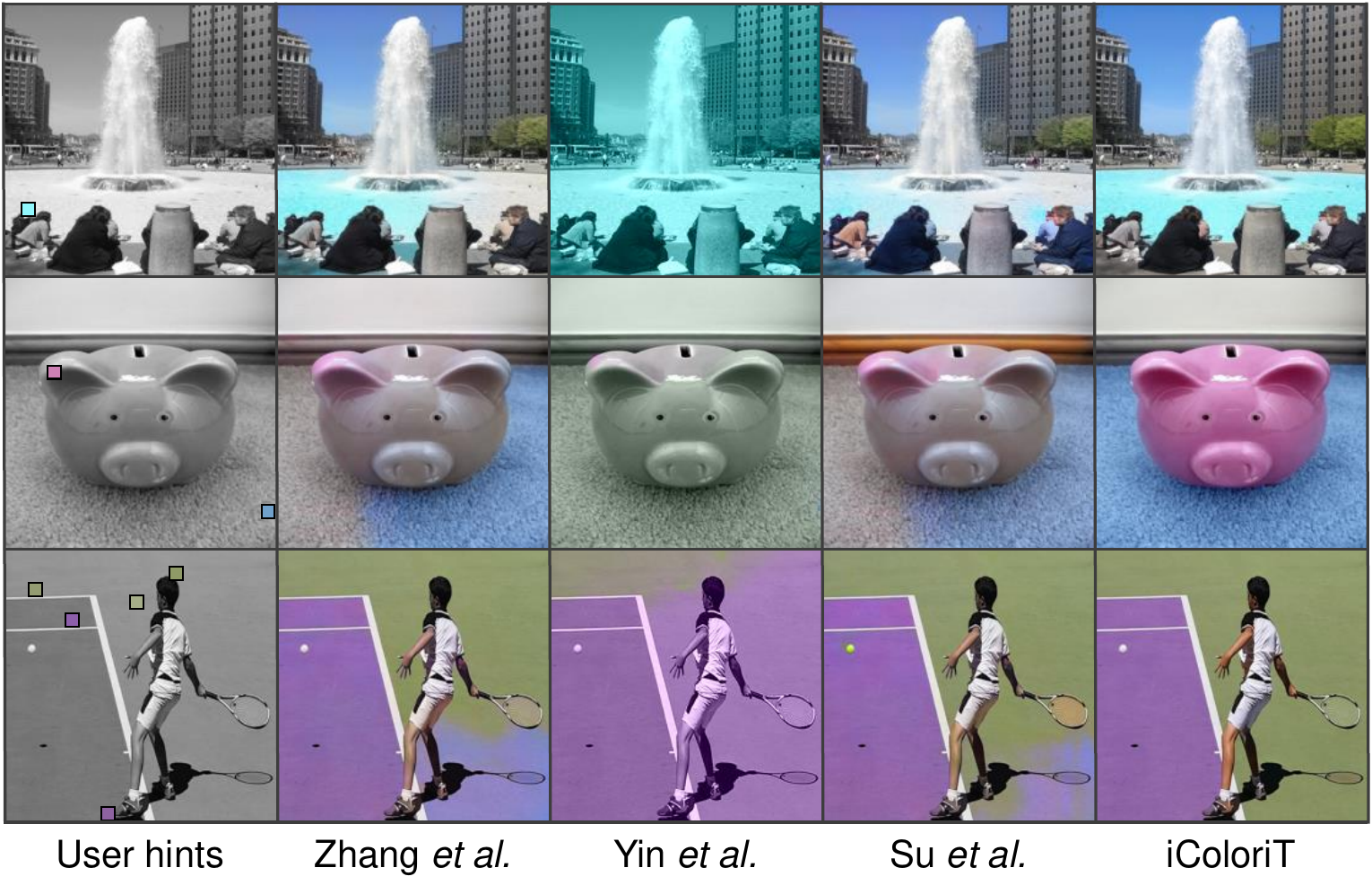}
    \caption{Example results of various point-interactive colorization approaches. Previous approaches often produce partially colorized results even where the grayscale values are persistent (\eg, water, floor, and grass), which indicates that the user hints did not properly propagate to the relevant regions.} 
    \label{fig:intro_quali}
    \vspace{-0.7cm}
\end{figure}

Practical point-interactive colorization methods assist the user to produce a colorized image with minimal user interaction. 
Thus, accurately estimating the regions relevant to the user hint can be beneficial for reducing the amount of user interactions. 
For example, using hand-crafted filters~\cite{levin2004, side} to determine the region a user hint should fill in was an early approach for colorizing simple patterns within the image. 
Recently, Zhang~\etal~\cite{zhang2017} proposed a learning-based model trained on a large-scale dataset~\cite{imagenet} which produces colorized images with a simple U-Net architecture. 
However, existing methods tend to suffer from partially colorized results even in obvious regions where the grayscale values are persistent, as seen in \Cref{fig:intro_quali}. 
This is due to the inefficient design of stacking convolutional layers in order to propagate hints to distant relevant regions. 
In other words, propagating hints to large semantic regions can only be done in the deep layers, which makes colorizing larger semantic regions more challenging than colorizing smaller regions. 
To overcome this hurdle, we leverage the global receptive field of self-attention layers~\cite{attention} in Vision Transformers~\cite{vit}, enabling the model to selectively propagate user hints to relevant regions at each single layer. 

Learning how to propagate user hints to other regions aligns well with the self-attention mechanism. 
Specifically, directly computing the similarities of features from all spatial locations (\ie, the similarity matrix) can be viewed as deciding where the hint colors should propagate in the entire image. 
Thus, in this work, we present iColoriT, a novel point-interactive colorization framework utilizing a modified Vision Transformer for colorizing grayscale images. 
To the best of our knowledge, this is the first work to employ a Vision Transformer for point-interactive colorization. 

Furthermore, promptly displaying the results for a newly provided user hint is essential for assisting users to progressively colorize images without delay. 
For this reason, we generate color images by leveraging the efficient pixel shuffling operation~\cite{pixelshuffle}, an upsampling technique that reshapes the output channel dimension into a spatial resolution. 
Through the light-weight pixel shuffling operation, we are able to discard the conventional decoder architecture and offer a faster inference speed compared to existing baselines. 
Despite its efficiency, pixel shuffling with large upsampling ratios tends to generate unrealistic images with missing details and notable boundaries as seen in \Cref{fig:intro_mae}. 
Therefore, we present the local stabilizing layer, which restricts the receptive field of the last layer, to mitigate the artifacts caused by pixel shuffling. 
Our contributions are as follows:
\begin{itemize}
    \item We are the first work to utilize a Vision Transformer for point-interactive colorization enabling users to selectively colorize relevant regions. 
    \item We achieve real-time colorization of images by effectively upsampling images with minimal cost, leveraging the pixel shuffling and the local stabilizing layer. 
    \item We provide quantitative and qualitative results demonstrating that iColoriT highly outperforms existing state-of-the-art baselines and generates reasonable results with fewer user interactions.
\end{itemize}

\vspace{-0.2cm}
\section{Related Work}
\label{sec:related_work}
\vspace{-0.2cm}

\vspace{+0.1cm}
\noindent \textbf{Interactive Colorization} 
Learning-based methods for image colorization ~\cite{coltran, chromagan, instanceaware, cic, Iizuka2016letcolor, ctest, memo, pixelated2020zhao, ctest} have proposed fully-automated colorization methods, which generate reasonable color images without the need of any user intervention. 
Interactive colorization methods~\cite{zhang2019deep, he2018deep, levin2004, zhang2017, side, lu2020gray2colornet, xu2020stylization, xiao2020example, yin2021yes, li2019automatic, li2021globally} are designed to colorize images given a user's condition which conveys color-related information. 
A widely-studied condition type for interactive colorization methods are reference images~\cite{he2018deep,li2019automatic, zhang2019deep, xiao2020example, lu2020gray2colornet, xu2020stylization, li2021globally, yin2021yes}, which are already-colored exemplar images. 
Using reference images can be convenient since the user can provide the overall color tones with a single image. 
However, it is difficult for the user to further edit specific regions in the colorized image since a new reference image is likely to produce a different colorization result. 

\begin{figure}[t]
    \centering
    \includegraphics[width=\columnwidth]{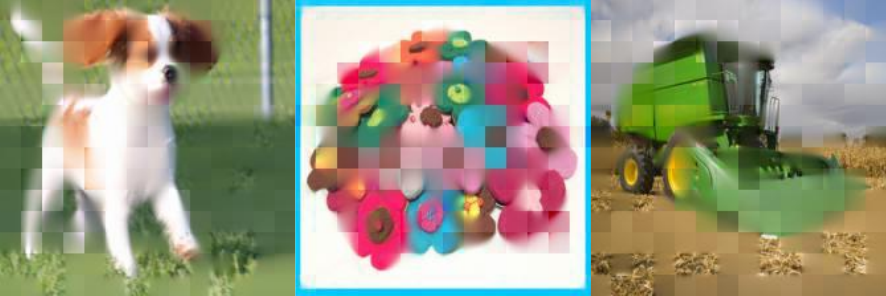}
    \caption{Images generated with large upsample ratios~\cite{mae} tends to suffer from evident borders between image patches. }
    \label{fig:intro_mae}
    \vspace{-0.5cm}
\end{figure}

\vspace{+0.1cm}
\noindent \textbf{Point-interactive Colorization} 
Point-interactive colorization models~\cite{levin2004, zhang2017, side} allow the user to progressively colorize images by specifying colors (\ie, user hints) at different point locations in the input grayscale image.
Since commonly used point sizes for specifying the spatial locations range from $2\times 2$ to $7\times 7$ pixels, the user hints only cover a small portion of the entire image.  
Thus, a point-interactive colorization model is required to propagate user hints to the entire image in order to produce a reasonable result with minimal user interaction. 
Early approaches~\cite{levin2004, side} utilized hand-crafted image filters to determine the propagation region of each hint by detecting simple patterns. 
The colors of the user hints are then propagated within each region using optimization techniques. 
Recently, Zhang~\etal~\cite{zhang2017} proposed a learning-based method by extending an existing unconditional colorization model~\cite{cic} to produce color images given a grayscale image and user hints. 
Although these methods use user hints as a condition for generating color images, common failure cases presented in \Cref{fig:intro_quali} indicate that the models often propagate hints incompletely. 
Stacking convolutional layers to propagate user hints indicates that propagating hints to distant relevant regions can only be done in the deeper layers, which makes colorizing larger semantic regions more challenging than nearby regions. 
Thus, we utilize the self-attention layer to enable user hints to propagate to any relevant regions at all layers. 

\begin{figure*}[ht]
    \centering
    \includegraphics[width=\textwidth]{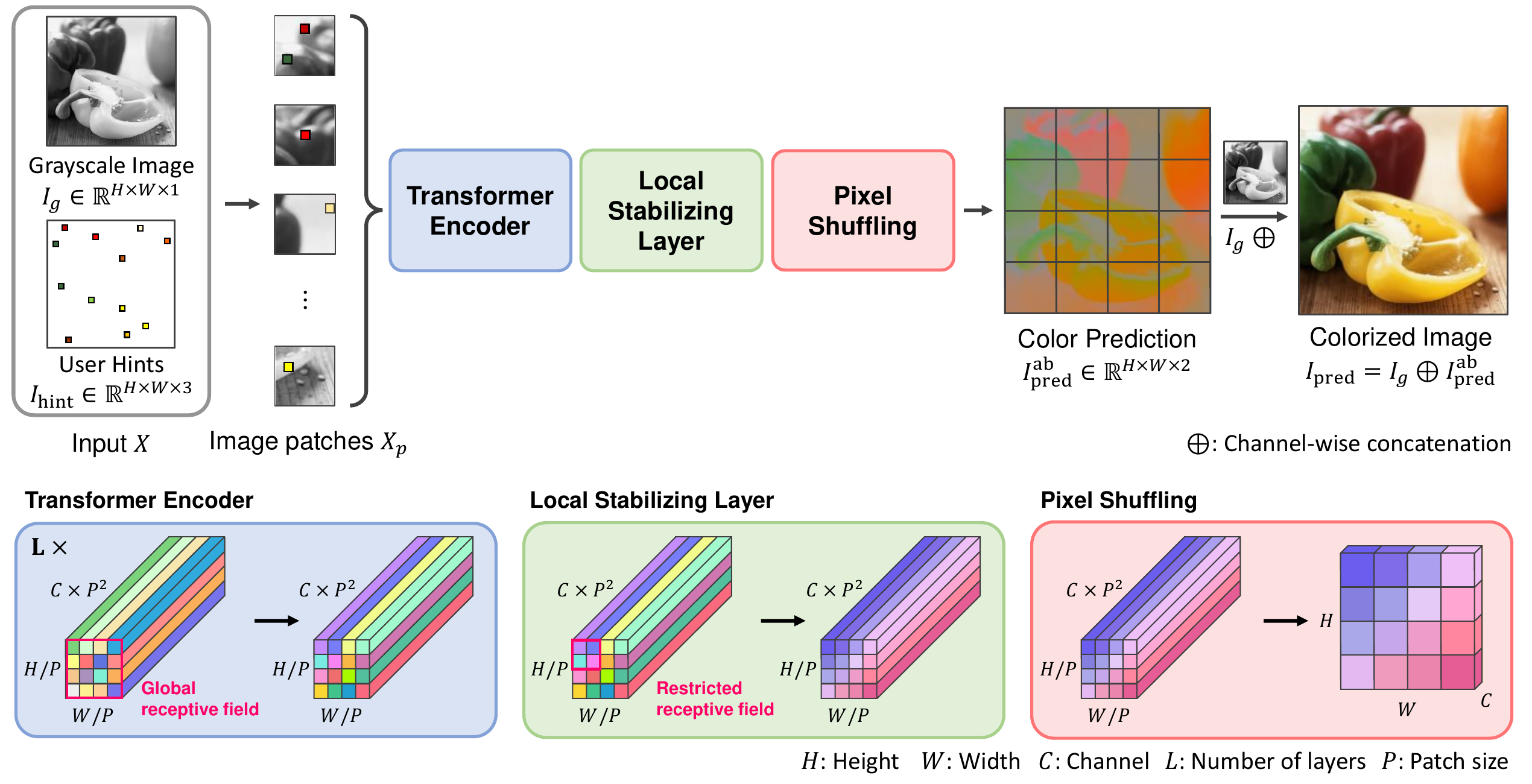}
    \vspace{-0.6cm}
    \caption{The overall workflow of iColoriT. We first obtain input $X$ by concatenating the grayscale image $I_g$ and the user hint $I_{\text{hint}}$ containing color conditions. The input is reshaped into input patches for the Transformer encoder. The output features from the Transformer encoder are passed through the local stabilizing layer and the pixel shuffling layer to obtain the final colors $I_{\text{pred}}^{\text{ab}}$. $I_{\text{pred}}^{\text{ab}}$ is then concatenated with $I_g$ to produce the colorized image.}
    \label{fig:method_main}
    \vspace{-0.5cm}
\end{figure*}

\vspace{+0.1cm}
\noindent \textbf{Image Colorization with Transformers} 
Unlike the widely-used convolution-based approach for image synthesis, recent studies~\cite{coltran, mae, simmim, vitgan} made efforts to synthesize images by only utilizing the Transformer architecture. 
Colorization Transformer (ColTran)~\cite{coltran} proposes an autoregressive model for unconditional colorization which uses the Transformer decoder architecture~\cite{attention} in order to generate diverse colorization results. 
Despite its outstanding performance for unconditional colorization, the excessively slow inference speed of autoregressive models hinders its application to user-interactive scenarios. 
Specifically, it takes 3.5-5 minutes to colorize a batch of 20 images of size $64\times 64$ images even with a P100 GPU. 
In this work, we leverage the Transformer \textit{encoder} to generate the colors of a grayscale image. 
The multi-head attention of the Transformer encoder enables our approach to generate color images with a single forward pass which reduces the inference time of our model compared to autoregressive colorization.

\noindent\textbf{Upsampling via Pixel Shuffling}
Pixel shuffling~\cite{pixelshuffle} is an upsampling operation that rearranges a $(H, W, C\times P^2)$ sized feature map into a shape of $(H\times P, W\times P, C)$ where each channel in the original feature map is reshaped into a $P\times P$ image patch. 
This can be viewed as \emph{upsampling via reshaping}, and is often used in super-resolution approaches to effectively upsample an image with minimal computational overhead. 
ViTGAN~\cite{vitgan} is a pioneering approach for using Transformer encoders for image generation and synthesizing image patches by pixel shuffling the output feature map. 
However, the usage of pixel shuffling was limited to a small upscale factor of 4 or 8, restricting the model to generate images in small resolutions (\eg, $32 \times 32 \text{ and } 64\times 64$). 
A known issue~\cite{mae} with pixel shuffling with larger upsampling ratios ($P > 8$) was that output images tend to contain evident borders between image patches as seen in \Cref{fig:intro_mae}. 
This is due to upsampling different image patches from different locations in the feature map. 
To overcome this hurdle, we present a local stabilizing layer, which promotes neighboring image patches to have coherent colors, allowing iColoriT to effectively upsample images to higher resolutions (\ie, $224\times 224$) without such artifacts. 

\vspace{-0.2cm}
\section{Proposed Method}

\vspace{-0.1cm}
\subsection{Preliminaries}

We first prepare the grayscale image $I_g\in \mathbb{R}^{H\times W \times 1}$ and the simulated user hints $I_{\text{hint}}\in \mathbb{R}^{H\times W \times 3}$ to be used as our training sample. 
A grayscale image $I_g$ can be acquired from large-scale datasets by converting the color space from RGB to CIELab~\cite{cie} and taking the L or lightness value. 
Similarly, the color condition $I_{\text{hint}}$ provided by the user can be expressed with the remaining a, b channel values $\Tilde{I}_{\text{hint}} \in \mathbb{R}^{H\times W \times 2}$ by filling the a,b channel values of all non-hint regions with 0. 
The user hint $I_{\text{hint}}\in \mathbb{R}^{H\times W \times 3}$ is constructed by adding a third channel to $\Tilde{I}_{\text{hint}}$ that marks hint regions with 1 and non-hint regions with 0. 

During training, we simulate the user hints by determining the hint location and the color of the hint. 
We sample hint locations from a uniform distribution since a user may provide hints anywhere in the image. 
Once the hint location is decided, the color of the user hint is obtained by calculating the average color values for each channel within the hint region since a user is expected to provide a single color for a single hint location. 
Finally, given the grayscale image $I_g\in \mathbb{R}^{H\times W \times 1}$ and the simulated user hints $I_{\text{hint}}\in \mathbb{R}^{H\times W \times 3}$, we obtain our input $X\in \mathbb{R}^{H\times W \times 4}$ by
\begin{equation*}
    X = I_g \oplus I_{\text{hint}},
\end{equation*}
where $\oplus$ is the channel-wise concatenation. 

\subsection{Propagating User Hints with Transformers}

We utilize the Vision Transformer~\cite{vit} to achieve a global receptive field for propagating user hints across the image as shown in \Cref{fig:method_main}. 
We first reshape our input $X\in \mathbb{R}^{H\times W \times 4}$ into a sequence of tokens $X_p \in \mathbb{R}^{N\times (P^2 \times 4)}$, where $H,W$ are the height and width of the original image, $P$ is the patch size, and $N=HW/P^2$ is the number of input tokens (\ie, sequence length). 
Thus, a $P\times P\times 4$ size image patch from the original input $X$ is used as a single input token.
These sequence of input tokens are passed through the Transformer encoder, which computes the input as, 
\begin{equation}
    z_0 = X_p + E_{pos}, \quad E_{pos} \in \mathbb{R}^{N\times d}
\end{equation}
\vspace{-0.4cm}
\begin{equation}
    z'_l = \text{MSA}(\text{LN}(z_{l-1})) + z_{l-1},
\end{equation}
\vspace{-0.3cm}
\begin{equation}
    z_l = \text{MLP}(\text{LN}(z'_l)) + z'_l,
\end{equation}
\vspace{-0.3cm}
\begin{equation}
    y_p = \text{LN}(z_L),
\end{equation}
where $E_{pos}$ denotes the sinusoidal positional encoding~\cite{vit}, $\text{MSA}(\cdot)$ indicates the multi-head self-attention~\cite{attention}, $\text{LN}(\cdot)$ indicates the layer normalization~\cite{layernorm}, $d$ denotes the hidden dimension, $l$ denotes the layer number, and $y_p \in \mathbb{R}^{N\times d}$ denotes the output of the Transformer encoder. 
Since self-attention does not utilize any position-related information, we add positional encoding $E_{pos}$ to the input and relative positional bias~\cite{swin, raffel2020exploring, hu2018relation, hu2019local} in the attention layer. 
Thus, the attention layer is computed as, 
\begin{equation}
    \text{Attention}(Q,K,V) = \text{softmax}(QK^T / \sqrt{d} + B)V, 
\label{eq:qk}
\end{equation}
where $Q,K,V \in \mathbb{R}^{N\times d}$ are the query, key and value matrices, $B \in \mathbb{R}^{N\times N}$ is the relative positional bias. 
The colors of the user hints are able to propagate to any spatial location at all layers due to the global receptive field of the self-attention mechanism.

\subsection{Pixel Shuffling and the Local Stabilizing Layer}
The output features of the Transformer encoder $y_p \in \mathbb{R}^{N\times d}$ can be viewed as a feature map $y \in \mathbb{R}^{H/P \times W/P \times d}$ of the original image. 
The spatial resolution of the output feature map $y$ is smaller than the resolution of the input image by a factor of $P$ since image patches of size $P\times P$ consists of a single input token. 
Therefore, the output feature map $y$ needs to be upsampled in order to obtain a full-resolution color image. 
While previous approaches~\cite{zhang2017, instanceaware} leverage a decoder for upsampling, we utilize pixel shuffling~\cite{pixelshuffle} which is an upsampling technique rearranging a ($H/P, W/P, C\times P^2$) feature map into a shape of ($H, W, C$) to obtain a full-resolution image. 

\begin{figure}[t]
    \centering
    \includegraphics[width=\columnwidth]{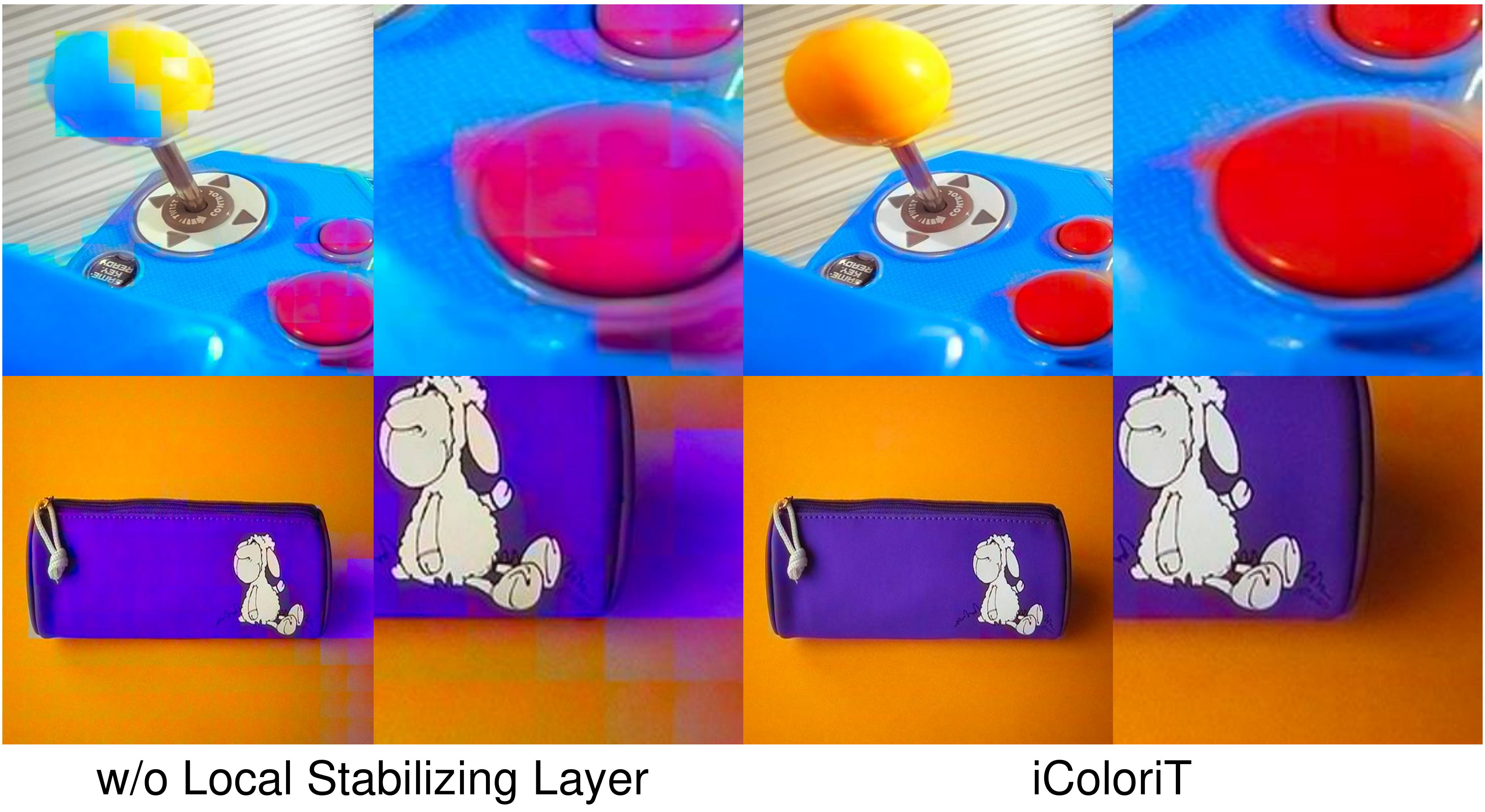}
    \vspace{-0.4cm}
    \caption{Example images of inconsistent colorization results observed in images produced without the local stabilizing layer. }
    \label{fig:ablation_conv}
    \vspace{-0.2cm}
\end{figure}

However, as mentioned in \Cref{sec:related_work}, large upsampling ratios (\eg, $P > 8$) may lead to images with visible artifacts along the image patch boundaries as seen in \Cref{fig:ablation_conv}. 
Thus, in order to promote reasonable generation of colors, we propose a local stabilizing layer, which restricts the model to generate colors utilizing neighboring features, and place the layer before pixel shuffling. 
We provide experiments in \Cref{sec:quali-ablation} with various design choices for the local stabilizing layer (\eg, linear, convolutional layer, and local attention) and select a simple yet effective convolutional layer as our final model. 
To sum up, our upsampling process can be written as, 
\begin{equation}
    I_{\text{pred}}^{\text{ab}} = \mathcal{PS}(\text{LS}(y)),
\end{equation}
where $\mathcal{PS}(\cdot)$ is the pixel shuffling operation, LS($\cdot$) is the local stabilizing layer, and $I_{\text{pred}}^{\text{ab}} \in \mathbb{R}^{H\times W \times 2}$ is the ab color channel outputs. 
The predicted color image $I_{\text{pred}} \in \mathbb{R}^{H\times W \times 3}$ is obtained by 
\begin{equation*}
    I_{\text{pred}} = I_g \oplus I_{\text{pred}}^{\text{ab}},
\end{equation*}
which is the concatenation of the given grayscale input $I_g$ (L channel)  and $I_{\text{pred}}^{\text{ab}}$ (ab channel). 
Through pixel shuffling and the local stabilizing layer, we can effectively obtain a full-resolution color image \emph{without} an additional decoder, allowing real-time colorization for the user (\Cref{sec:quanti}). 

\subsection{Objective Function}

We train our model with the Huber loss~\cite{huber} between the predicted image and the original color image in the CIELab color space, 
\vspace{-0.2cm}
\begin{equation}
\begin{split}
    L_{recon} = &\frac{1}{2}(I_{\text{pred}} - I_{GT})^2\mathbbm{1}_{\left|I_{\text{pred}} - I_{GT}\right| < 1} \\ 
    &+ (\left|I_{\text{pred}} - I_{GT}\right|-\frac{1}{2})\mathbbm{1}_{\left|I_{\text{pred}} - I_{GT}\right| \geq 1}.
\end{split}
\vspace{-0.2cm}
\end{equation}

\begin{figure*}[ht]
    \centering
    \includegraphics[width=\linewidth]{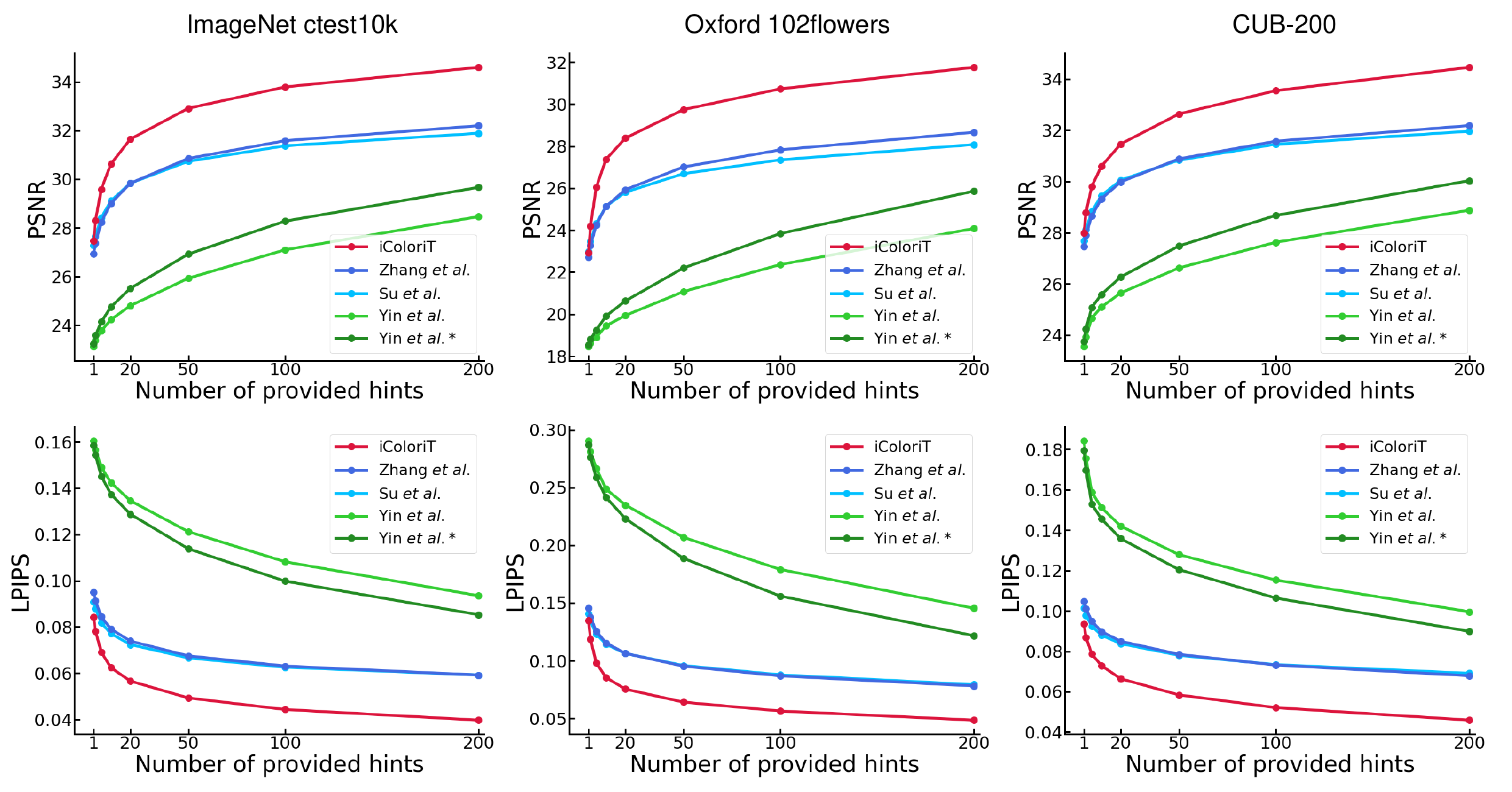}
    \vspace{-0.6cm}
    \caption{Average PSNR and LPIPS of the test images according to the number of provided hints. Hint locations are sampled from a uniform distribution and $2\times 2$ hints are revealed to the model. Yin~\etal$^*$~\cite{side} denotes the results evaluated with $2\times 2$ hints and Yin~\etal~\cite{side} denotes the results evaluated with $7\times 7$ hints. iColoriT outperforms existing approaches by a large margin as the number of provided hints increases.}
    \label{fig:exp_quanti_psnr}
    \vspace{-0.5cm}
\end{figure*}

\section{Experiments}
\label{sec:impl_details}
\vspace{-0.1cm}

\noindent \textbf{Implementation Details} 
We follow the configurations of ViT-B~\cite{vit} for the Transformer encoder blocks. 
For the local stabilizing layer, we use a single layer with a receptive field of 3. 
We experiment with two types of layers (\Cref{sec:quali-ablation}), the local attention and the convolutional layer, and use the simple yet effective convolutional layer as the default local stabilizing layer. 
For training, we resize images to a $224\times 224$ resolution and use a patch size of $P=16$ which also becomes the upsampling ratio. 
Thus, the sequence length $N$ is 196 and the last output dimension $d$ is 512. 
We sample hint locations uniformly across the image and sample the number of hints from a uniform distribution $\mathcal{U}(0, 128)$. 
We provide experiments on different model sizes, patch sizes, the local stabilizing layer, and the number of hints in \Cref{sec:quali-ablation} and the supplementary material. 

We use the AdamW optimizer~\cite{adamw} with a learning rate of $0.0005$ managed by the cosine annealing scheduler~\cite{coslr}. 
The model is trained for 2.5M iterations with a batch size of 512. 

\vspace{+0.1cm}
\noindent \textbf{Datasets}
For training, we use the ImageNet 2012 train split~\cite{imagenet} which consists of 1,281,167 images. 
We do not use the classification labels during training since our model is trained in a self-supervised manner. 
We evaluate our method on three datasets from different domains, all of which are colorful validation datasets suitable for evaluating colorization approaches. 
Note that we do not additionally finetune the model for each validation dataset. 
The ImageNet ctest10k10k~\cite{ctest} is a subset of the ImageNet validation split used as a standard benchmark for evaluating colorization models. 
ImageNet ctest10k excludes any grayscale image from ImageNet and consists of 10,000 color images. 
We also evaluate on the Oxford 102flowers dataset~\cite{flowers} and the CUB-200 dataset~\cite{cub} which provide 1,020 colorful flower images from 102 categories and 3,033 samples of bird images from 200 different species, respectively. 

\noindent \textbf{Baselines}
We compare the performance of iColoriT with existing interactive colorization methods~\cite{zhang2017, side}. 
We also extend a recent unconditional colorization model by Su~\etal~\cite{instanceaware}, which utilizes an off-the-shelf object detector~\cite{maskrcnn} to individually color multiple instances, to a point-interactive colorization model. 
Since the model proposed by Su~\etal~\cite{instanceaware} employs the same model architecture and objective function as the point-interactive colorization model by Zhang~\etal~\cite{zhang2017}, we are able to effortlessly extend the approach to a point-interactive colorization method by conditioning the model with user hints in the same manner. 
The extended model is trained under the configurations provided by Zhang~\etal~\cite{zhang2017} and Su~\etal~\cite{instanceaware} using ImageNet~\cite{imagenet}. 
Note that although the model proposed by Su~\etal~\cite{instanceaware} is trained with the ImageNet~\cite{imagenet} dataset, this approach is assisted by an off-the-shelf object detector pre-trained on a large-scale object detection dataset~\cite{coco}. 
All baselines are trained and evaluated with the publicly available official codes.

\begin{figure*}[ht]
    \centering
    \includegraphics[width=\linewidth]{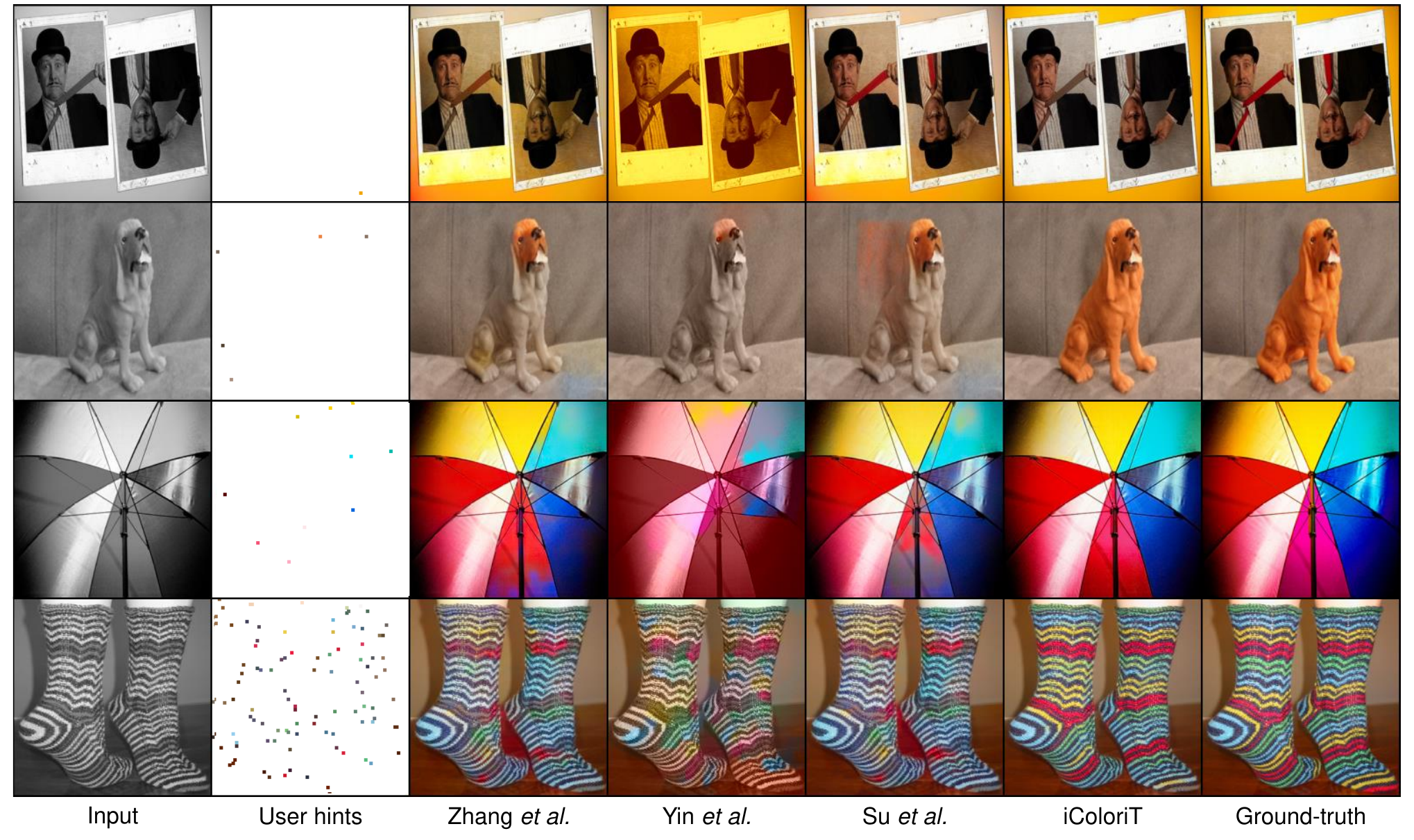}
    \caption{Qualitative results of point-interactive colorization methods given 1, 5, 10, and 100 user hints. iColoriT is able to produce reasonable color images by appropriately propagating user hints. }
    \label{fig:exp_quali}
    \vspace{-0.5cm}
\end{figure*}

\vspace{-0.1cm}
\subsection{Comparison with Existing Approaches}
\label{sec:quanti}
\vspace{-0.1cm}

\noindent \textbf{Quantitative Evaluation of iColoriT}
We plot the average peak signal-to-noise ratio (PSNR) and the learned perceptual image patch similarity (LPIPS)~\cite{lpips} of the test images according to the number of provided hints in \Cref{fig:exp_quanti_psnr}. 
For evaluating the point-interactive colorization models, we simulate user hints with the ground-truth colors from the image, considering a situation where the user intends to colorize the grayscale image into the original color image. 
User hints are simulated by randomly selecting hint locations from a uniform distribution. 
The hint sizes are set to $2\times 2$ and the hint color is given as the average color within each hint region in the original color image following the protocol of Zhang~\etal~\cite{zhang2017}. 
We empirically find that smaller hint sizes are usually beneficial for both the colorization model and the user in terms of receiving and giving accurate color conditions. 
However, the method proposed by Yin~\etal~\cite{side} assumes that a user provides an abundant amount of user hints. 
Thus, we further evaluate this method by revealing larger hints of size $7\times 7$ which is the result we report for all following evaluations. 

We empirically find that methods proposed by Zhang~\etal~\cite{zhang2017} and Su~\etal~\cite{instanceaware} tend to arbitrarily colorize images without reflecting user hints.
While this may be helpful for achieving a relatively higher initial PSNR when the arbitrarily colorized color is the ground-truth color, it hinders further control for the user to achieve a high PSNR in subsequent stages of colorization. 
As seen in \Cref{fig:exp_quanti_psnr}, iColoriT quickly reflects the user hints and aids the user to efficiently colorize grayscale images with minimal interaction. 
The PSNR in the early stages of colorization notably increases with each additional hint. 
The results indicate that iColoriT highly outperforms existing baselines for generating colorized images a user specifically has in mind.

\begin{figure}[t]
    \centering
    \includegraphics[width=0.95\columnwidth]{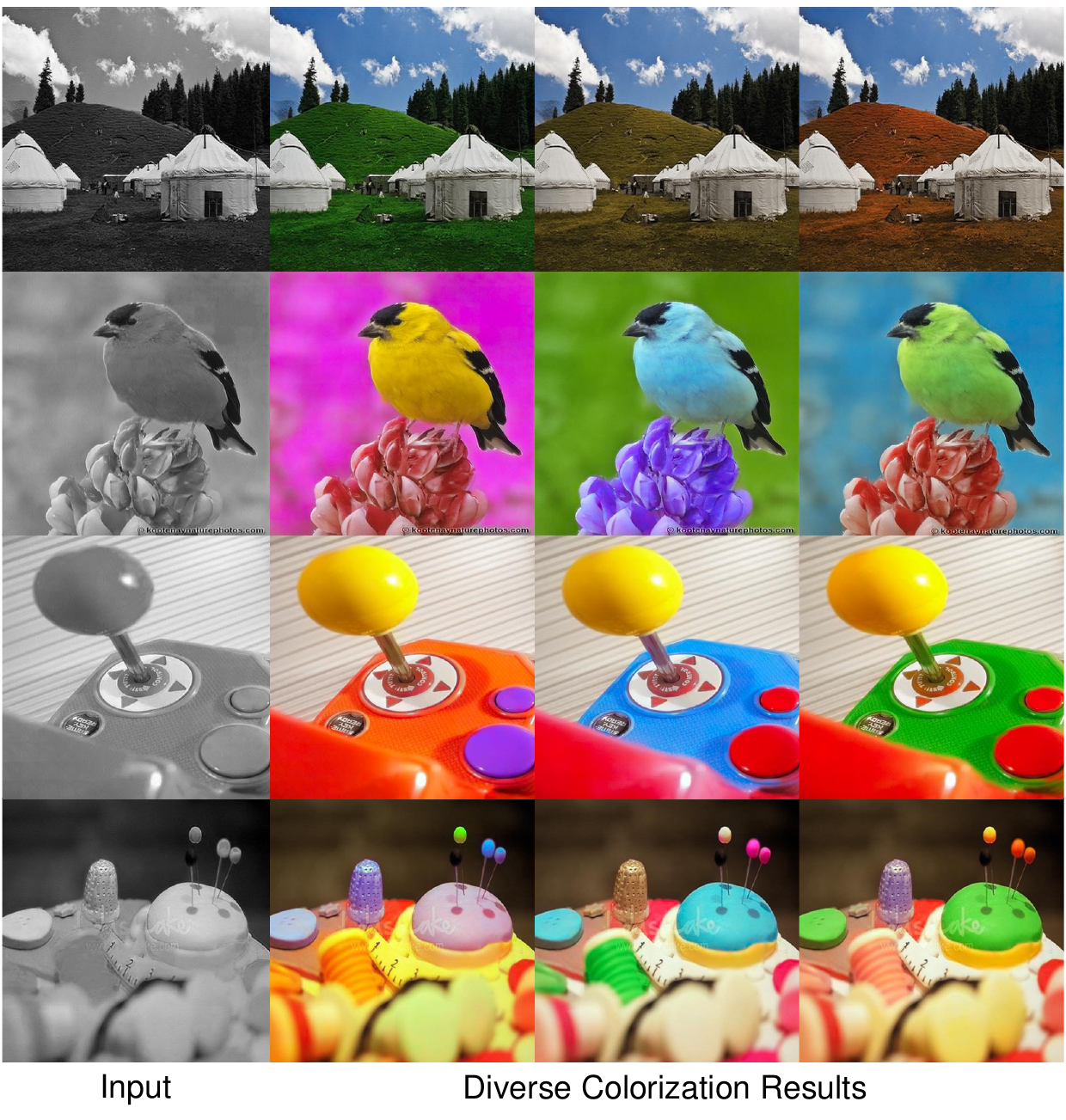}
    \caption{Images colorized with different colors provided by the user. The images from the ImageNet ctest10k~\cite{ctest} are colored by hand-picking hint locations and changing the hint colors. }
    \label{fig:exp_multi_color}
    \vspace{-0.2cm}
\end{figure}

\noindent \textbf{Qualitative Results of iColoriT}
\label{sec:quali}
We provide qualitative results produced by the baselines and iColoriT in \Cref{fig:exp_quali} when given an original grayscale image and the simulated user hints. 
iColoriT is able to produce realistic images that closely resemble the ground-truth image indicating that a user can colorize images as they please. 
Also, as seen in the colorized results in \Cref{fig:intro_quali} and \Cref{fig:exp_quali}, iColoriT is capable of appropriately colorizing large areas even with a small number of user hints while other approaches leave most regions uncolored or incorrectly colored. 
iColoriT can also colorize detailed regions when given a sufficient number of hints as shown in the last row of \Cref{fig:exp_quali}.

iColoriT is also suitable for producing \emph{diverse} colorized images when given various user hints as seen in \Cref{fig:exp_multi_color}. 
Instead of the simulated user hints from the ground-truth image, we provide multiple sets of hand-picked user hints to colorize a single grayscale image. 
We fix the hint locations for an image and alter the user-provided colors to observe the colorized results. 
iColoriT can produce various realistic colorization results that reflect the intention of the user.
We provide uncurated qualitative results and a demo video in the supplementary material. 
Also, we will release the iColoriT demo including the graphical user interface, providing a powerful tool for image colorization. 

\begin{table}[t]
\centering
\begin{tabular}{@{}x{2.1cm}x{1.8cm}x{1.9cm}@{}}
\toprule
Methods    & PSNR@10        & LPIPS@10 \\ \midrule
iColoriT-T & 28.86          & 0.084    \\
iColoriT-S & 29.67          & 0.073    \\ \midrule
iColoriT   & \textbf{30.63} & \textbf{0.062} \\ \bottomrule
\end{tabular}
\vspace{0.1cm}
\caption{Scalability of iColoriT to lightweight models. PSNR and LPIPS given 10 user hints (PSNR@10 and LPIPS@10) on the ImageNet ctest10k~\cite{ctest} are reported for each model. }
\label{tab:exp_smaller_scale}
\vspace{-0.5cm}
\end{table}

\noindent\textbf{Scaling to Lightweight Models}
iColoriT can easily scale to smaller models and still achieve high performance. 
We train iColoriT in smaller scales using the configurations of the ViT-S and the ViT-Ti~\cite{vit2} for our Transformer encoder. 
We report the PSNR and the LPIPS given 10 hints (PNSR@10 and LPIPS@10) for ImageNet ctest10k and compare them against other models in \Cref{tab:exp_smaller_scale}. 
We were able to train iColoriT-S and iColoriT-T with only a slight performance drop and still maintain a high performance. 
We believe that the Transformer architecture and the self-attention mechanism are central for propagating hints to larger semantic regions, achieving a high PSNR even in small-scale models.

\noindent \textbf{Real-time Inference}
The inference speed (\ie, latency) of point-interactive models is important for providing a satisfying user experience. 
Thus, we measure the time required for a single forward pass and compare it with the latency of baseline models in \Cref{tab:exp_latency}. 
We report the speed on both CPU and GPU using a commercial AMD Ryzen 5 PRO 4650G and a single NVIDIA RTX 3090. 
We also provide the number of floating-point operations (FLOPs) and the number of parameters required for each model. 
We were not able to measure GPU latency, FLOPs, and the number of parameters for Yin~\etal~\cite{side} since the method is not a learning-based model. 
The model proposed by Su~\etal~\cite{instanceaware} operates in two stages, an initial object detection stage and an instance-wise colorization stage. 
We only report the latency for the second stage which still exhibits a slow inference speed since the colorization model needs to color multiple objects individually. 
Due to the efficient pixel shuffling for upsampling images, iColoriT enjoys a short latency of 540ms and 14ms on a CPU and GPU device respectively, providing real-time colorization results for the user. 
iColoriT-T and iColoriT-S show an exceptionally fast inference speed on a CPU-only device (\ie, 177ms and 253ms, respectively), which makes the model an appealing option when considering applications to real-world scenarios where accelerators may not be available.

\begin{table}[t]
\centering
\begin{tabular}{@{}cR{1.5cm}R{1.5cm}R{1.6cm}@{}}
\toprule
Methods                 & \begin{tabular}[c]{@{}c@{}}CPU\\Latency\end{tabular} & \begin{tabular}[c]{@{}c@{}}GPU\\Latency\end{tabular} & GFLOPs \\ \midrule
Zhang~\etal~\cite{zhang2017}
                        & 881ms & 24ms & 58.04 \\
Yin~\etal~\cite{side}
                        & 15,248ms & - & - \\
Su~\etal~\cite{instanceaware}
                        & 1,389ms & 45ms & 123.48 \\
iColoriT-T              & \textbf{177ms} & \textbf{13ms} & \textbf{1.43} \\
iColoriT-S              & 253ms & 14ms & 4.95 \\ \midrule
iColoriT                & 540ms & 14ms & 18.22 \\ \bottomrule
\end{tabular}
\vspace{0.1cm}
\caption{Inference speed of iColoriT and each baseline model. We provide the latency of each model in a CPU device and a GPU device along with the computational cost measured in FLOPs and number of parameters. }
\label{tab:exp_latency}
\vspace{-0.4cm}
\end{table}

\subsection{Ablation Study}
\vspace{-0.1cm}
\noindent \textbf{Designing the Local Stabilizing Layer}
\label{sec:quali-ablation}
We provide an ablation study on the local stabilizing layer by replacing it with different operations such as the linear layer and the local self-attention layer~\cite{localattn}. 
Using a linear layer can be viewed as eliminating the local stabilizing layer since a linear layer does not utilize neighboring features for generating the final output. 
In order to quantify the inconsistent color generation among image patches seen in \Cref{fig:ablation_conv}, we measure the mean squared error (MSE) for each image patch and report the variance of the errors within an image. 
We denote this measure the patch error variance (PEV). 
A high PEV implies that the model has varying accuracy depending on the image patch. 
The local stabilizing layer resolves this issue in a simple yet effective manner by predicting the ab channel values of an image patch from neighboring output features as illustrated in \Cref{fig:method_main}. 
We also measure the PSNR near the image patch boundaries (\ie, one pixel from the patch borders) to observe the accuracy in the regions containing inconsistent color generation. 
As seen in \Cref{tab:exp_ablation_conv}, adding an operation with a limited receptive field (\ie, convolution and local self-attention) lowers the PEV and increases the PSNR along the patch boundaries, indicating that the model generates colors with consistent accuracy across the image.
The convolutional layer serves as a simple yet effective approach for reducing artifacts caused by pixel shuffling and generating realistic colorized images. 

\begin{table}[t]
\centering
\begin{tabular}{@{}x{2.4cm}x{1.6cm}x{1.95cm}x{1.0cm}@{}}
\toprule
Methods               & PSNR@10        & B-PSNR@10            & PEV$\downarrow$ \\ \midrule
Linear                & 28.78          & 28.71                & 39.39            \\
Local Attention       & 28.85          & 28.77                & 38.82            \\ \midrule
Convolution           & \textbf{28.86} & \textbf{28.80}       & \textbf{38.81}   \\ \bottomrule
\end{tabular}
\vspace{0.1cm}
\caption{Ablation study on the local stabilizing layer. PSNR@10, PSNR along the boundary (B-PSNR@10), and PEV on the ImageNet ctest10k~\cite{ctest} are reported for each model. All models are trained with the iColoriT-T configuration.}
\label{tab:exp_ablation_conv}
\vspace{-0.3cm}
\end{table}

\begin{table}[t]
\centering
\begin{tabular}{@{}x{2.0cm}x{2.3cm}x{2.5cm}@{}}
\toprule
Patch Size              & PSNR@10               & CPU Latency \\ \midrule
$8 \times 8$            & 29.17\win{($+$0.31)}  & 373ms\lose{($+$196ms)} \\
$32 \times 32$          & 28.32\lose{($-$0.54)} & 147ms\win{($-$30ms)}   \\ \midrule
$16\times 16$           & 28.86                 & 177ms  \\ \bottomrule
\end{tabular}
\vspace{0.1cm}
\caption{iColoriT different upsampling ratios. PSNR@10, LPIPS@10, and CPU latency are reported for each model on the ImageNet ctest10k~\cite{ctest} test set. All models are trained with the iColoriT-T configuration.}
\label{tab:exp_ablation_patch}
\vspace{-0.5cm}
\end{table}

\noindent \textbf{Changing the Upsampling Ratio}
We experiment on various patch sizes $P$ (\ie, $P=8, 16,$ and $32$), which also becomes the upscaling ratio for pixel shuffling. 
While smaller patch sizes may allow fine-grained calculation of the similarity matrix, the computational cost escalates \emph{biquadratically}, since the computational complexity for the self-attention follows $\mathcal{O}(N^2)$ and $N = HW/P^2$ is the sequence length. 
Thus, we were not able to train our base model with a smaller patch size due to the prohibitive computational overhead. 
Instead, we compare the results on the smaller iColoriT-T model and report the average PSNR@10 and CPU latency on \Cref{tab:exp_ablation_patch}. 
While using a smaller patch size may be beneficial for achieving a higher PSNR, the increased computational cost hinders scaling to larger models for an additional performance gain and increases the CPU latency. 
We choose a patch size of $16\times 16$ since it can obtain both a short latency and a high PSNR while also being scalable to larger models (\ie, iColoriT-S and iColoriT).

\subsection{Visualizing the Internal Representation}
\begin{figure}[t]
    \centering
    \includegraphics[width=\columnwidth]{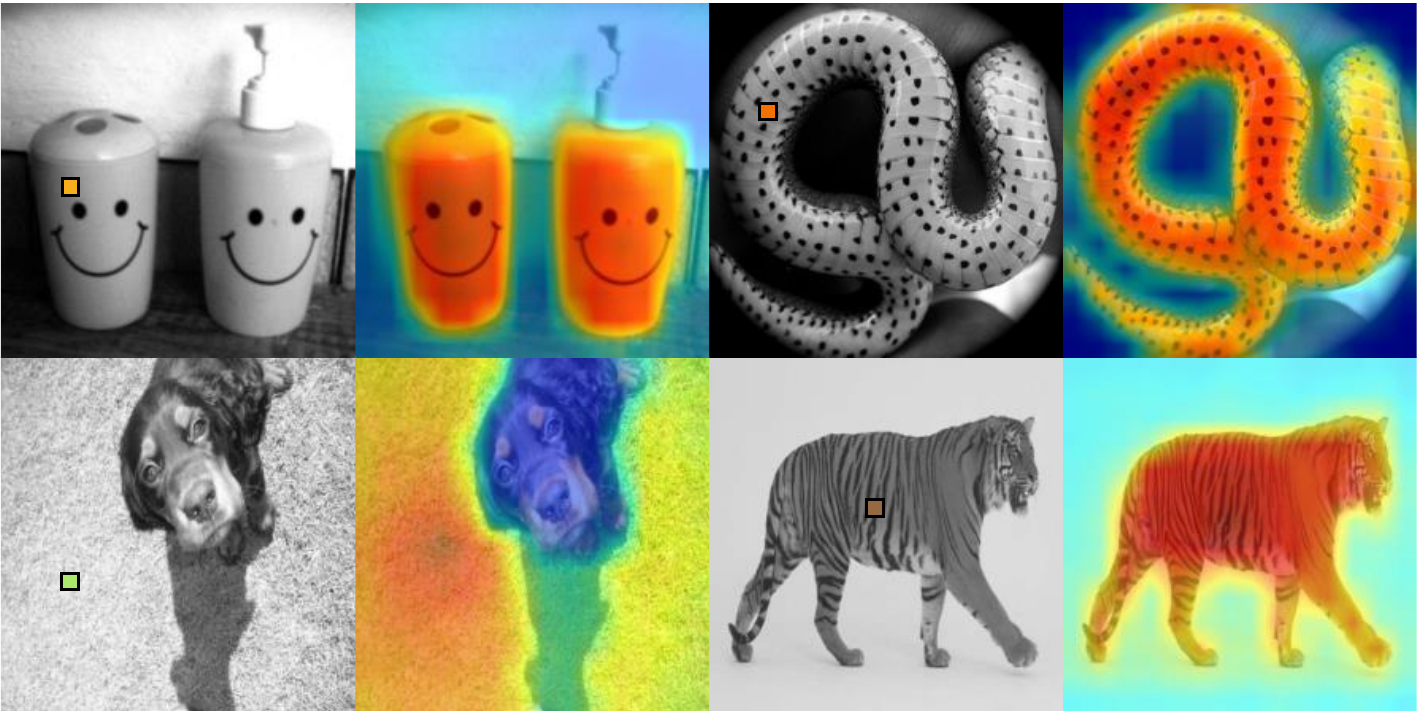}
    \vspace{-0.4cm}
    \caption{Visualization of the self-attention mechanism employing the attention rollout~\cite{rollout} method. iColoriT appropriately attends the user hint to relevant locations even for complex structures. }
    \label{fig:exp_rollout}
    \vspace{-0.32cm}
\end{figure}

\vspace{-0.1cm}
\label{sec:quali-matrix}
We further provide analysis on the self-attention mechanism to examine how our model is propagating user hints to other regions. 
We use the attention rollout method~\cite{rollout} to interpret the attention weights from the Transformer encoder for specific spatial locations. 
We visualize the attention maps for the input tokens which contain a user hint in \Cref{fig:exp_rollout}. 
Attention maps for hint locations can be directly interpreted as how the hint is propagating to other locations since tokens with high similarities are likely to be colorized with similar color as the color of the user hint. 
The self-attention mechanism enables iColoriT to selectively colorize relevant locations, even for regions with spatially complicated structures. 
These visualization aligns well with our qualitative and quantitative results demonstrating that iColoriT can effectively aid users to colorize images with minimal interaction.

\begin{figure}[t]
    \centering
    \includegraphics[width=\columnwidth]{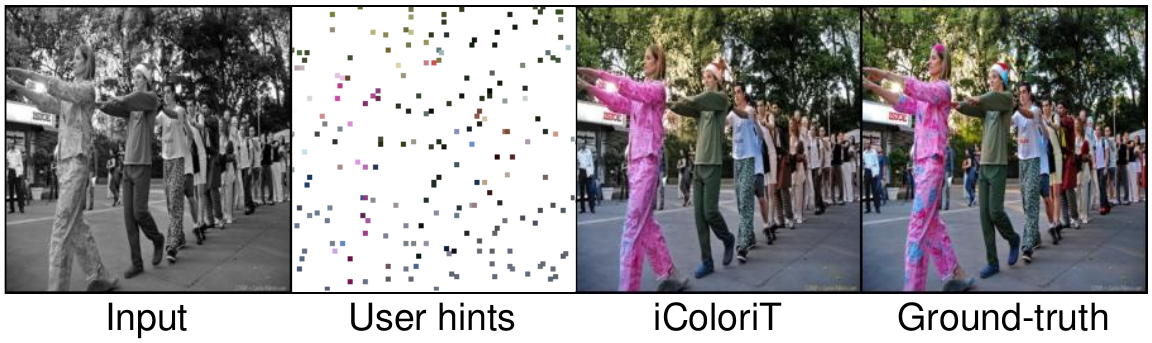}
    \vspace{-0.5cm}
    \caption{A common failure case for point-interactive colorization models in detailed regions.}
    \label{fig:exp_failure}
    \vspace{-0.6cm}
\end{figure}

\vspace{-0.2cm}
\section{Conclusion and Limitations}
\vspace{-0.2cm}

In this paper, we present iColoriT, a novel real-time point-interactive colorization framework capable of selectively propagating colors of the user hints to relevant regions. 
Through the Transformer encoder, pixel shuffling and the local stabilizing layer, iColoriT highly outperforms existing baselines, being able colorize images with minimal user interaction. 
Also, qualitative results indicate that iColoriT can generate diverse and realistic results when given various user hints. 
We justify our novel design through extensive experiments and ablation studies. 

Although iColoriT shows its strength even in detailed regions as shown in both quantitative and qualitative results, iColoriT may not be able to colorize small objects or distinguish close objects with the same grayscale intensity, since it does not leverage any semantic labels. 
This is a common drawback of point-interactive colorization approaches as seen in \Cref{fig:exp_failure} since models are trained in a self-supervised manner. 
Directly utilizing segmentation labels for training a point-interactive colorization model can be a promising future work. 
Nonetheless, we believe that the iColoriT is a practical application for real-world scenarios, effectively assisting the user to colorize images.

{\small
\bibliographystyle{ieee_fullname}
\bibliography{egbib}
}

\clearpage

\noindent \textbf{\huge{{Supplementary Materials}}}
\vspace{0.5cm}

This supplementary material presents quantitative results on unconditional colorization (\Cref{sec:uncon_fid}), the effect of the number of hints sampled during training (\Cref{sec:num_hint}), quantitative analysis of the hint propagation range (\Cref{sec:hpr}), and additional qualitative results (\Cref{sec:sup_quali}). 
We also attach a demo video demonstrating the use case of iColoriT.

\section{Unconditional Colorization Using iColoriT}
\label{sec:uncon_fid}
Similar to recent point-interactive colorization approaches~\cite{zhang2017, instanceaware}, iColoriT is capable of generating colorized images without any user hints. 
We compare the Fr\'{e}chet inception distance (FID) score~\cite{fid} of the colorization results generated without any user hints with a pre-trained Inception V3~\cite{inception} network. 
We exclude the conventional filter-based approach by Yin~\etal~\cite{side} since the method does not operate without user hints. 
The FID score is widely used to evaluate the how realistic the generated images are compared to the original color images. 
Our approach achieves a low FID score and generates realistic colors even without user hints. 

\begin{table}[ht]
\centering
\begin{tabular}{@{}x{2.4cm}x{1.8cm}@{}}
\toprule
Methods                       & FID           \\ \midrule
Zhang~\etal~\cite{zhang2017}  & 7.29          \\
Su~\etal~\cite{instanceaware} & 6.56          \\
Yin~\etal~\cite{side}         & -             \\ \midrule
iColoriT                      & \textbf{4.89} \\ \bottomrule
\end{tabular}
\vspace{0.1cm}
\caption{Fr\'{e}chet inception distance (FID) score of unconditional results of iColoriT and baselines. iColoriT is able to generate realistic colors even without user hints.}
\label{tab:supp_exp_fid}
\vspace{-0.5cm}
\end{table}


\section{Number of Hints Sampled During Training}
\label{sec:num_hint}

\begin{figure}[t]
    \centering
    \includegraphics[width=0.95\columnwidth]{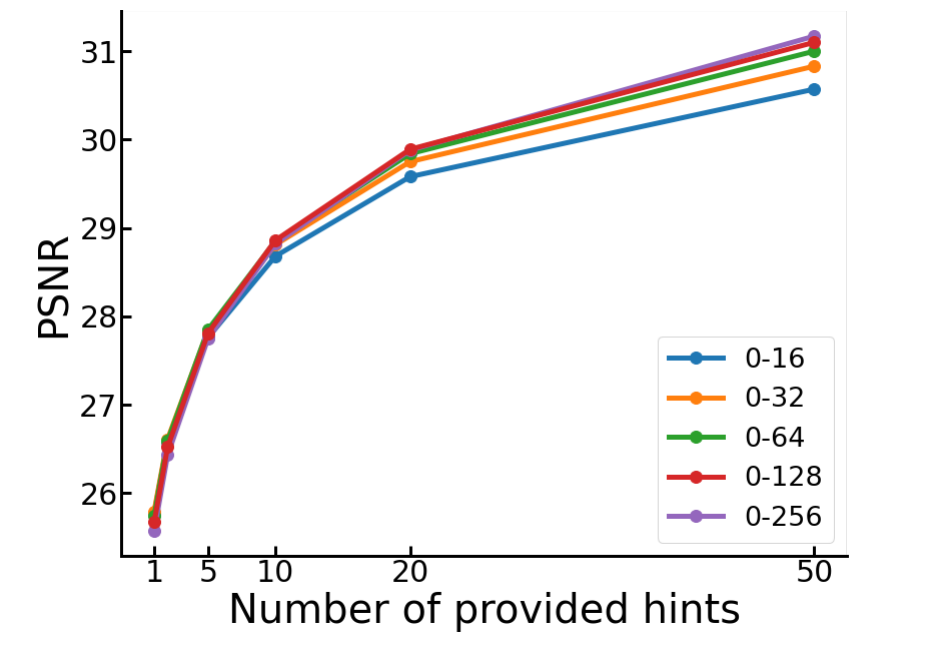}
    \caption{PSNR achieved by models trained by sampling different number of simulated user hints. The numbers on the legend indicate the range of the uniform distribution. The PSNR is measured in the ImageNet ctest10k~\cite{ctest} validation set.}
    \label{fig:supp_num_hints}
\end{figure}

The number of simulated user hints sampled during training may alter the performance since iColoriT directly learns how to propagate the color hints. 
We train our model provided with different number of hints sampled from various uniform distributions (\eg, $\mathcal{U}(0,16)$, $\mathcal{U}(0,32)$, $\mathcal{U}(0,64)$, $\mathcal{U}(0,128)$, $\mathcal{U}(0,256)$). 
To our surprise, the number of hints provided during training did not have an immense effect on the final performance. 
As plotted in \cref{fig:supp_num_hints}, the PSNR measured on ImageNet ctest10k~\cite{ctest} are similar for most models. 
We empirically find that sampling the number of hints from $\mathcal{U}(0,128)$ demonstrates a high performance in most regions including PSNR given 10 user hints (PSNR@10) which we choose as a representative indicator. 
We also observe that the performance of models given a relatively small number of hints (\ie, $\mathcal{U}(0, 16)$) does not necessarily outperform other models when evaluated with a small number of hints (\eg, PSNR@5). 
We presume that the model learns to appropriately propagate hints to relevant regions when trained with a more diverse number of hints. 
Future work focusing on how to sample the simulated hints is an interesting subject. 

\begin{figure*}[t]
    \centering
    \includegraphics[width=\textwidth]{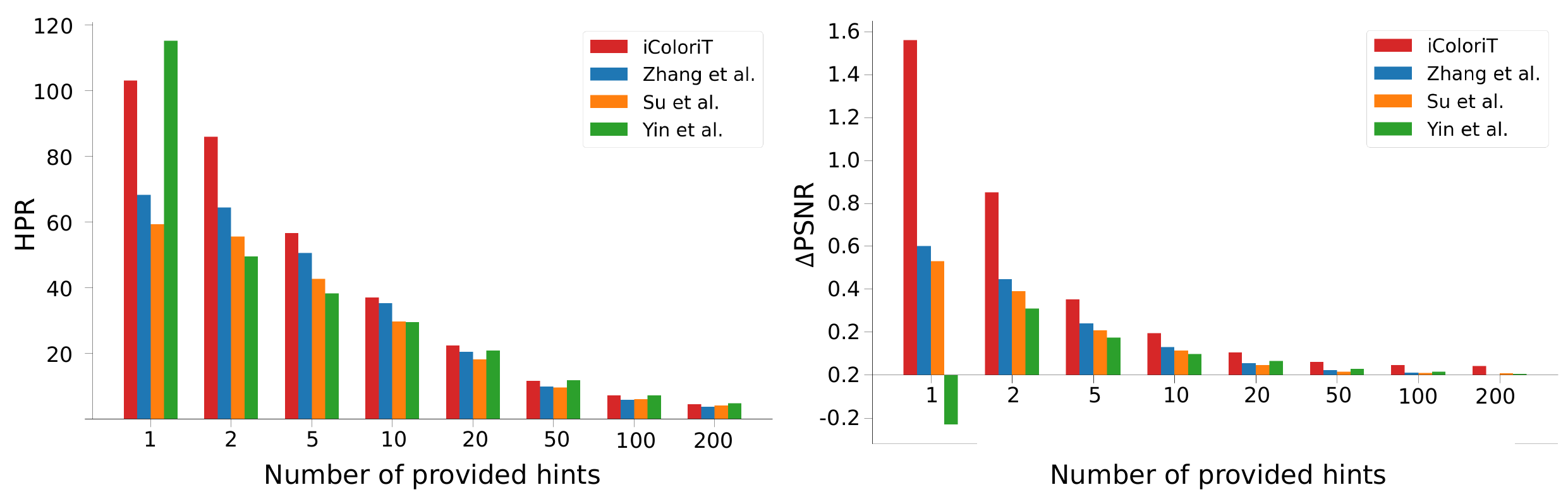}
    \vspace{-0.45cm}
    \caption{Average hint propagation range (HPR) and PSNR gain ($\Delta$PSNR) when given an additional hint. All scores are measured in the ImageNet ctest10k~\cite{ctest} dataset. iColoriT shows both high HPR and a high PSNR gain at all stages of the colorization process. }
    \label{fig:exp_quanti_hpr}
    \vspace{-0.2cm}
\end{figure*}

\begin{figure}[t]
    \centering
    \includegraphics[width=\columnwidth]{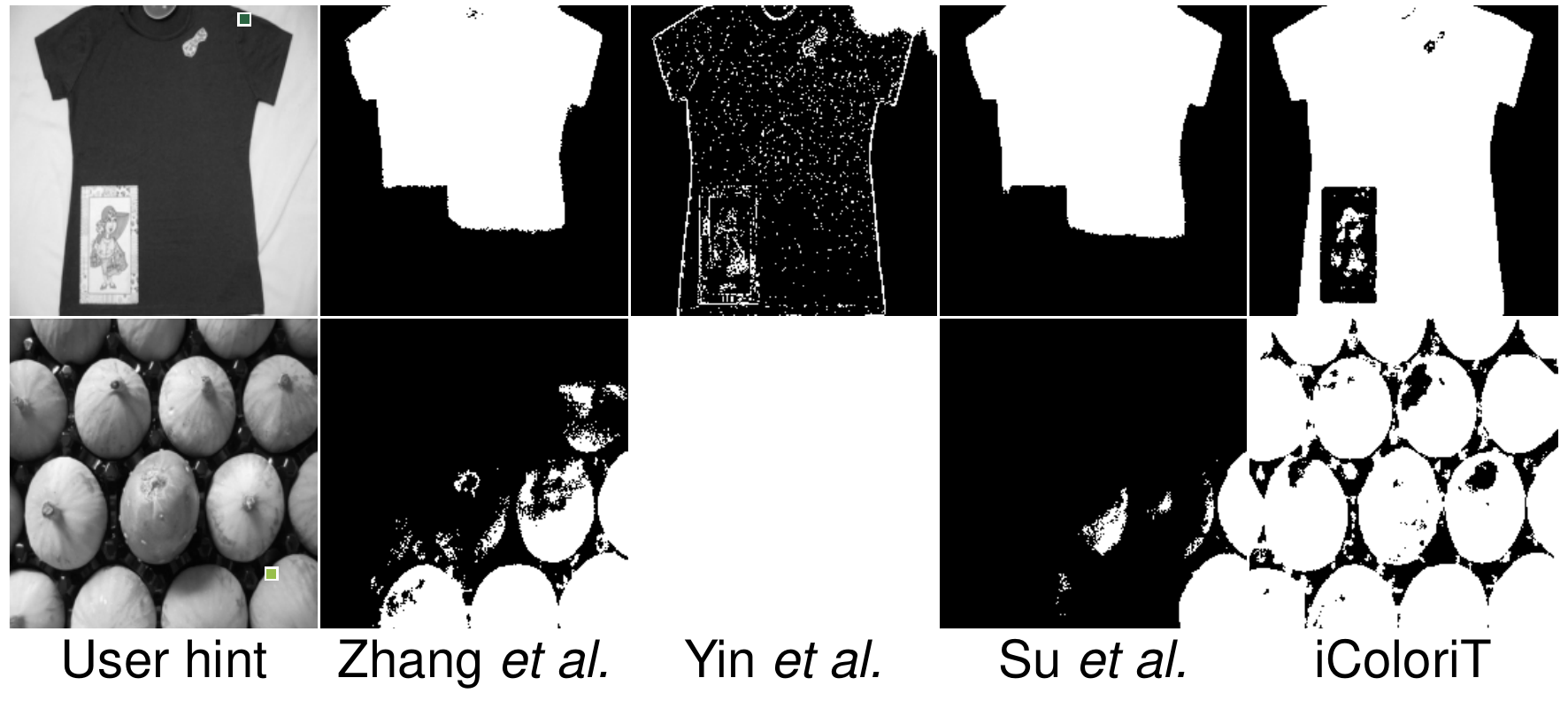}
    \vspace{-0.5cm}
    \caption{Visualization of the colorized region $\mathcal{C}_1$ for different baselines. iColoriT is able to propagate the hint to further regions if needed, while other approaches do not necessarily colorize all relevant regions. }
    \label{fig:exp_hpr_quali}
    \vspace{-0.5cm}
\end{figure}

\vspace{+0.1cm}
\section{Measuring the Hint Propagation Range}
\label{sec:hpr} 
Not only can iColoriT accurately reflect user hints and achieve a higher PSNR, iColoriT is capable of propagating user hints to longer distances if needed. 
In order to measure how far a hint propagates to further regions, we present the hint propagation range (HPR) measure. 
Given an image $I_{\text{pred}}^{t-1}$ colorized with $t-1$ number of color hints, HPR@~$t$ indicates the average distance from the newly provided $t$-th hint to the pixels that have been colorized by the $t$-th hint. 
We define a pixel to have been \textit{colorized} if the mean squared error (MSE) between the initial value and the altered value is larger than 2.3 in the CIELab color space~\cite{cie}, which is the just-noticeable-difference (JND) perceived by the human eye~\cite{jnd}. 
Given a set of coordinates $(x_i,y_i) \in \mathcal{C}_t$ of pixels colorized by hint $h_t$ and the coordinate of $h_t$ $(x_h, y_h)$, we calculate HPR with, 
\begin{equation}
    \mathcal{C}_t = \{(x, y)\;| \; \text{MSE}(I^t_{x y}, I^{t-1}_{x y}) > \text{JND} \},
\end{equation}
\begin{equation}
    \text{HPR}@\,t = \frac{1}{|\mathcal{C}_t|}\sum_{(x_i, y_i)\in \mathcal{C}_t} \sqrt{(x_i - x_h)^2 + (y_i-y_h)^2},
\end{equation}
which is the average Euclidean distance from $h_t$ to all locations in $\mathcal{C}_t$.

We measure the HPR across ImageNet ctest10k~\cite{ctest} at different stages of the colorization process and plot the results in \Cref{fig:exp_quanti_hpr}. 
Note that the HPR measure itself does not assess whether a model is appropriately reflecting the color hints provided by the user. 
However, when examined together with PSNR gain ($\Delta$PSNR) in \Cref{fig:exp_quanti_hpr} where the PSNR hugely improves at earlier stages, we can conclude that iColoriT reflects the color hints to further regions in a constructive manner. 
The optimization-based method proposed by Yin~\etal~\cite{side} often overly propagates the initial color hint to the entire image, which does not contribute to improving the PSNR or the perceptual quality compared to the original grayscale image. 
Learning-based baselines~\cite{zhang2017, instanceaware} on the other hand, tend to locally colorize images which hinders further PSNR gain. 
iColoriT takes advantage of both aspects and propagates color hints to further regions while hugely improving the PSNR. 

 \begin{figure}[t]
    \centering
    \includegraphics[width=0.65\columnwidth]{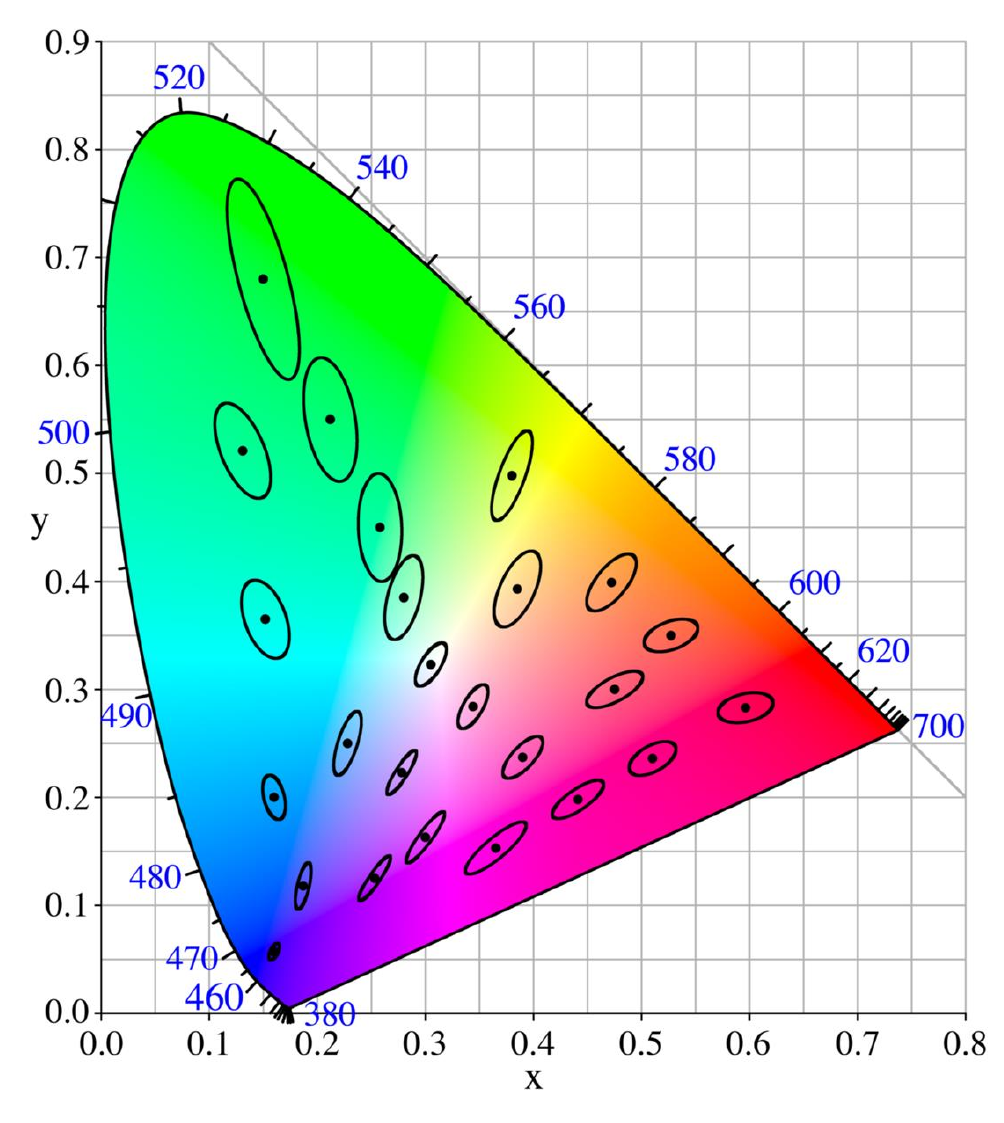}
    \caption{The MacAdam ellipse from MacAdam's paper~\cite{macadam} illustrating the just-noticeable-difference in the \textit{xy} chromaticity diagram.}
    \vspace{-0.5cm}
 \end{figure}

We also visualize $\mathcal{C}_1$, the pixel coordinates colorized by the initial hint $h_1$, in \Cref{fig:exp_hpr_quali} to intuitively understand the regions of which the hint alters the image. 
We believe that the self-attention mechanism is central for selectively colorizing the relevant regions regardless of the distance from the user-provided color hint. 
Qualitative results supporting this claim are provided in \Cref{sec:sup_quali}. 

For a clearer understanding, we provide an illustration of the MacAdam ellipse which visualizes the just-noticeable-difference of colors. 
The colors within the same ellipse indicate that the colors are indistinguishable to the human eye.
Although the areas of the ellipses are not uniform in the \textit{xy} chromaticity diagram, the JND in the CIELab color space is known to be roughly constant. 

\vspace{-0.2cm}
\section{Additional Qualitative Results}
\label{sec:sup_quali}

\begin{figure*}[t]
    \centering
    \includegraphics[width=0.85\textwidth]{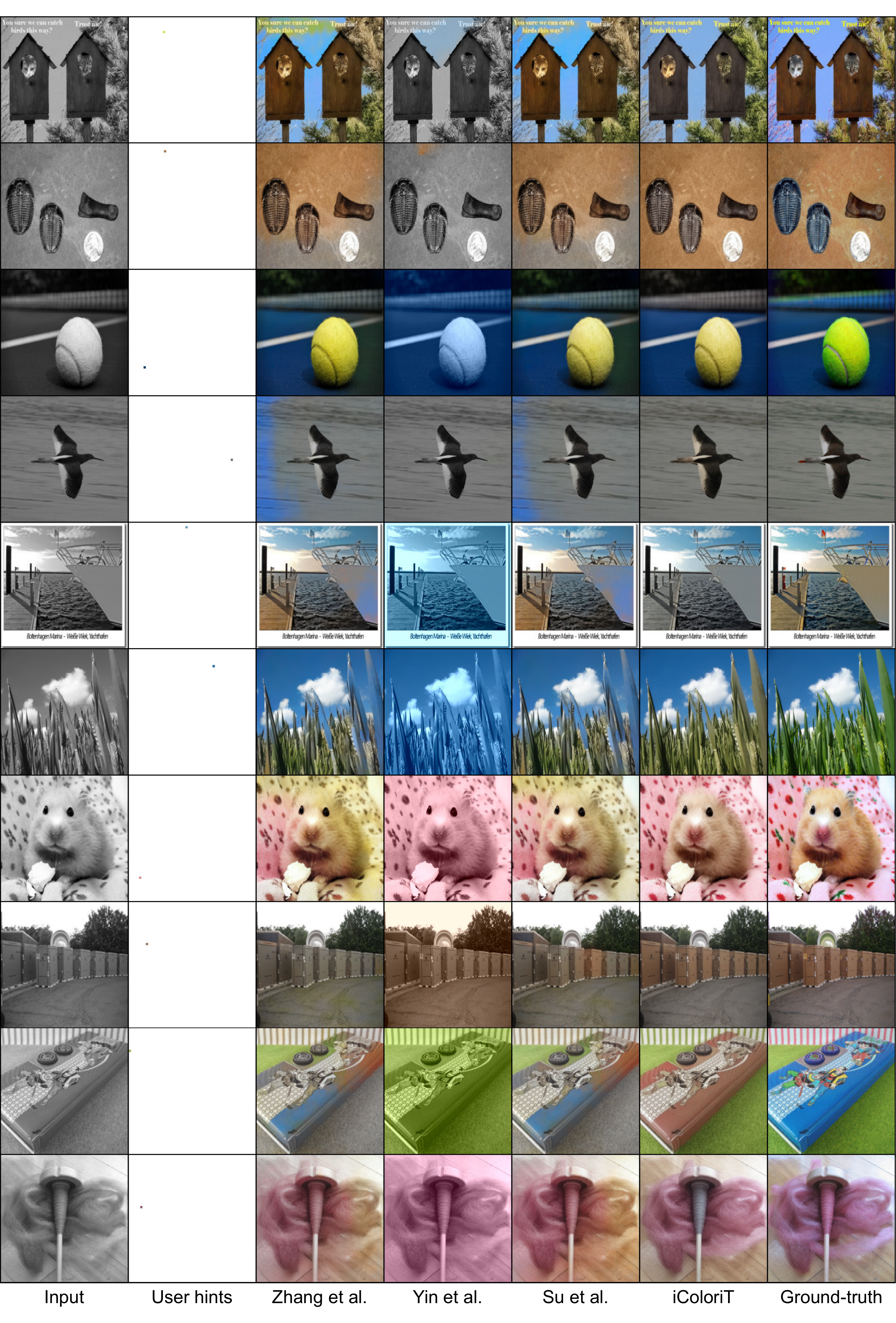}
    \caption{Additional qualitative results compared with baseline approaches. A single hint location is sampled from a uniform distribution. }
    \label{fig:add_quali_p1}
\end{figure*}

\begin{figure*}[t]
    \centering
    \includegraphics[width=0.85\textwidth]{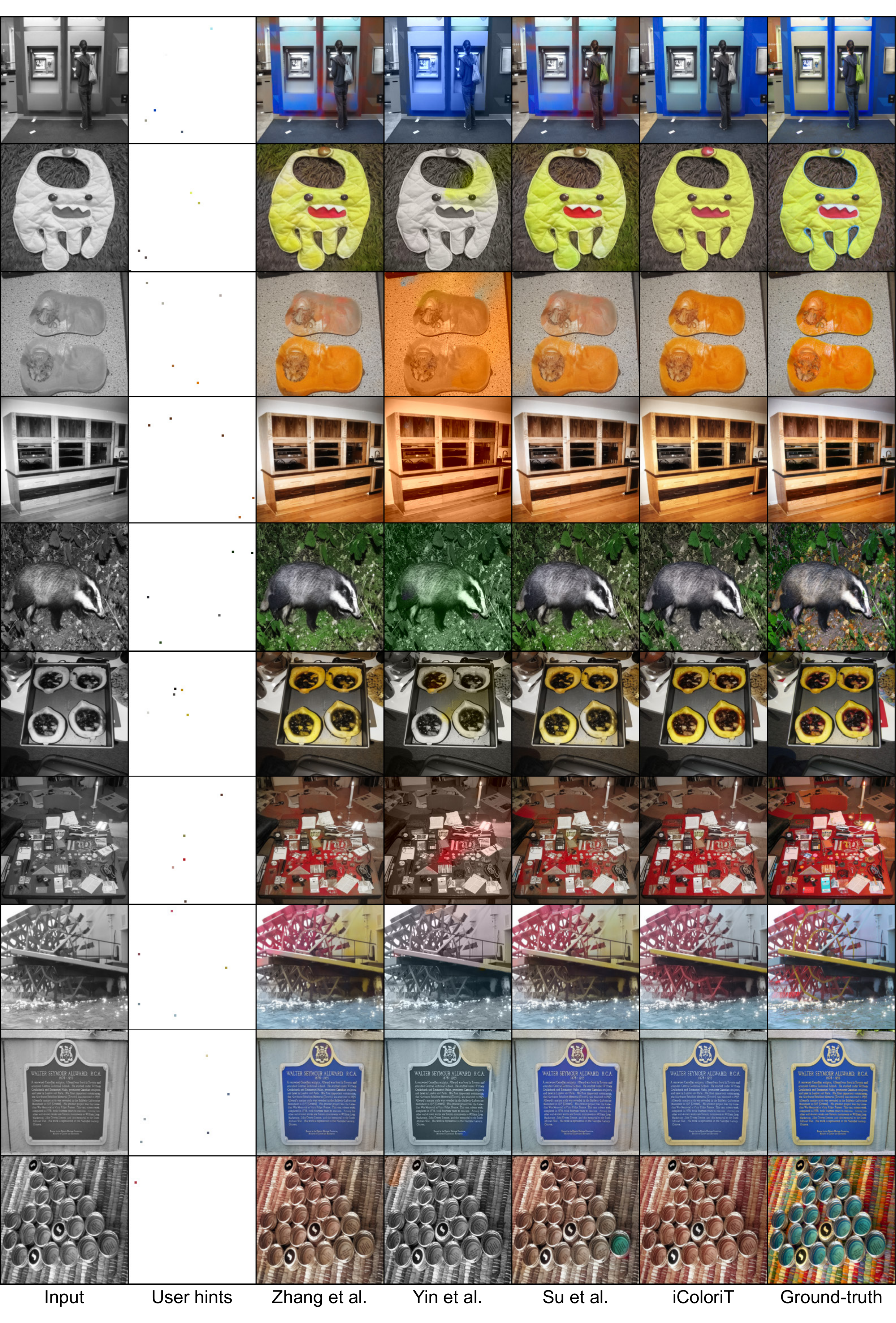}
    \caption{Additional qualitative results compared with baseline approaches. 5 hint locations are sampled from a uniform distribution. }
    \label{fig:add_quali_p5}
\end{figure*}

\begin{figure*}[t]
    \centering
    \includegraphics[width=0.85\textwidth]{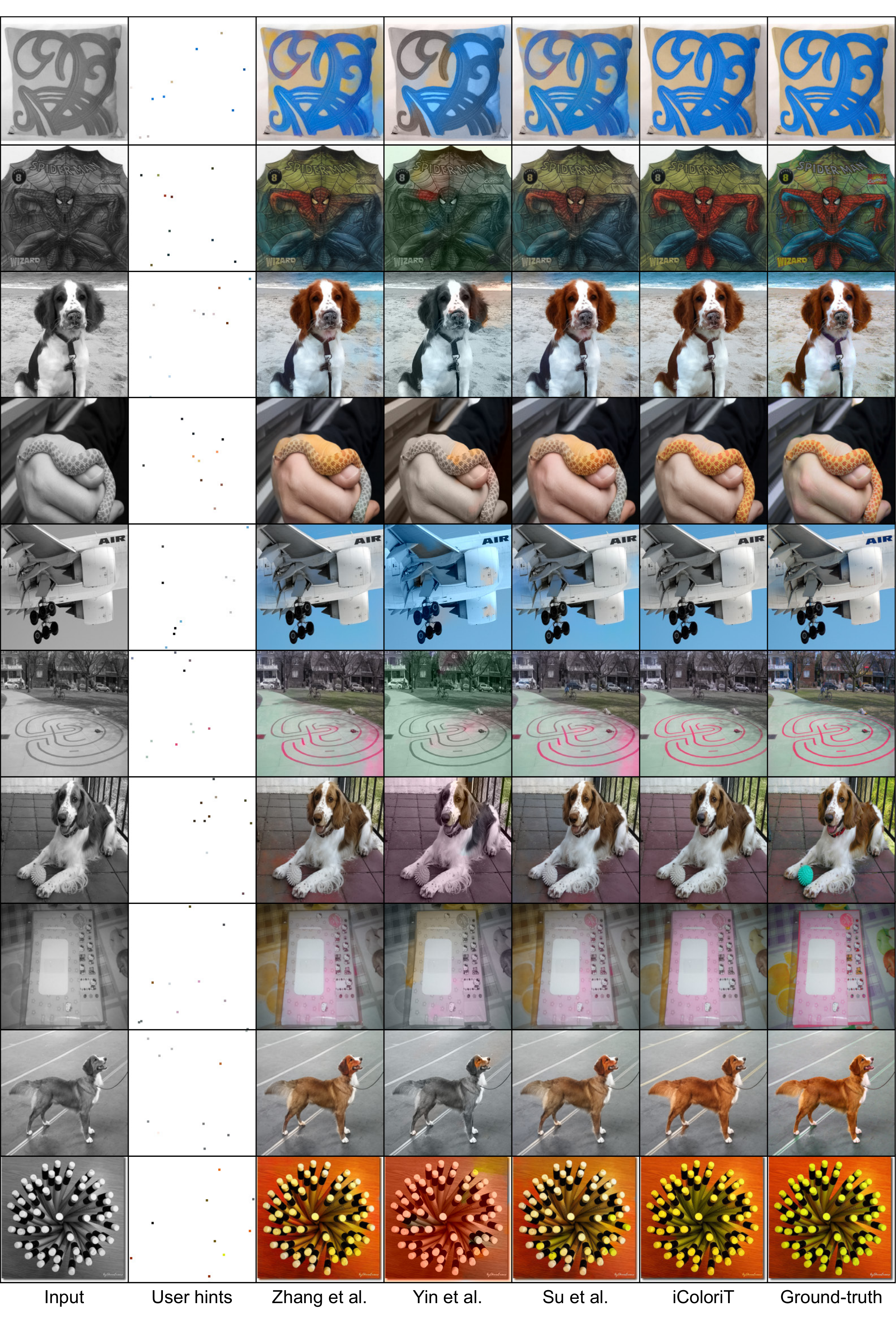}
    \caption{Additional qualitative results compared with baseline approaches. 10 hint locations are sampled from a uniform distribution. }
    \label{fig:add_quali_p10}
\end{figure*}

\begin{figure*}[t]
    \centering
    \includegraphics[width=0.85\textwidth]{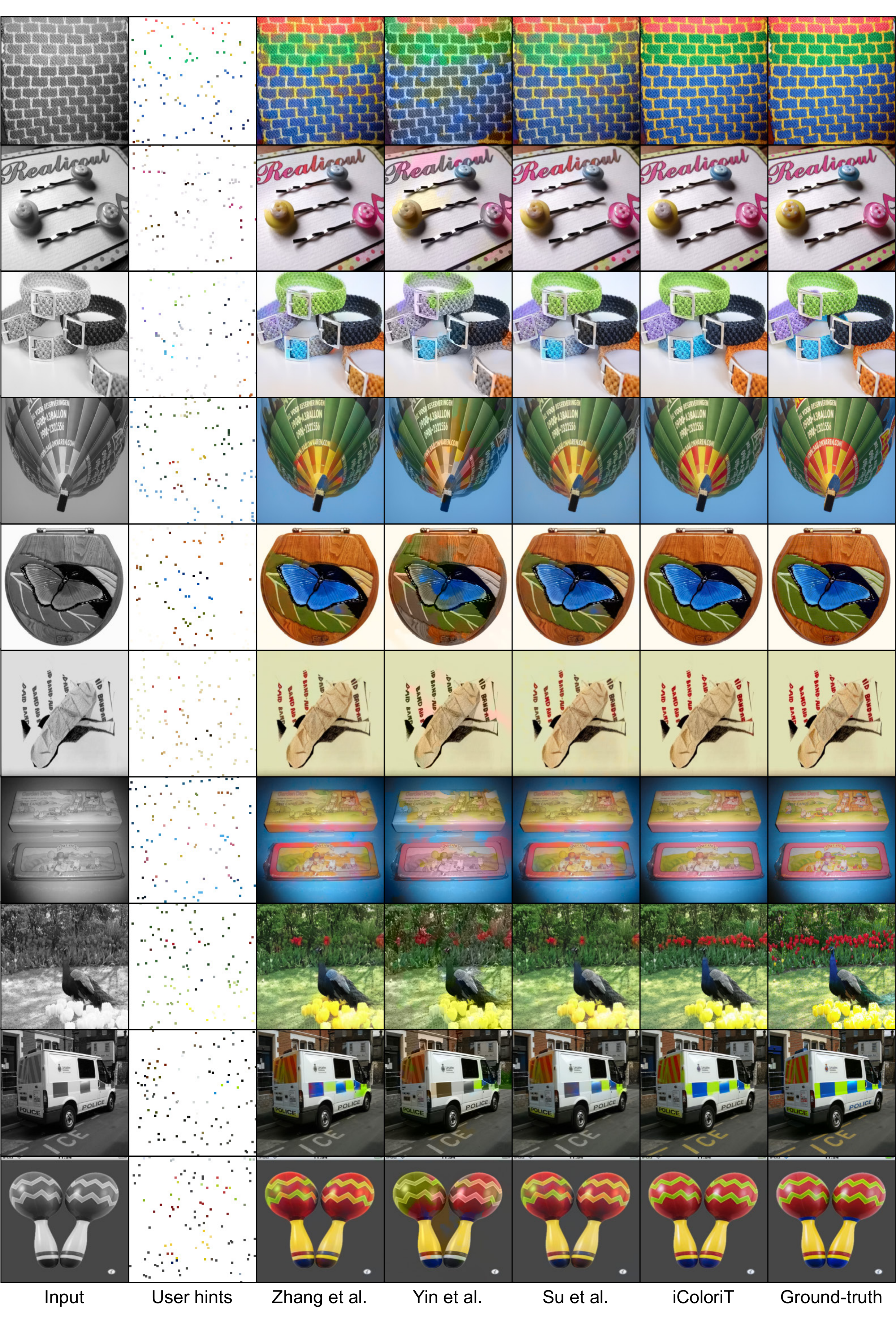}
    \caption{Additional qualitative results compared with baseline approaches. 100 hint locations are sampled from a uniform distribution. }
    \label{fig:add_quali_p100}
\end{figure*}

\begin{figure*}[t]
    \centering
    \includegraphics[width=0.92\textwidth]{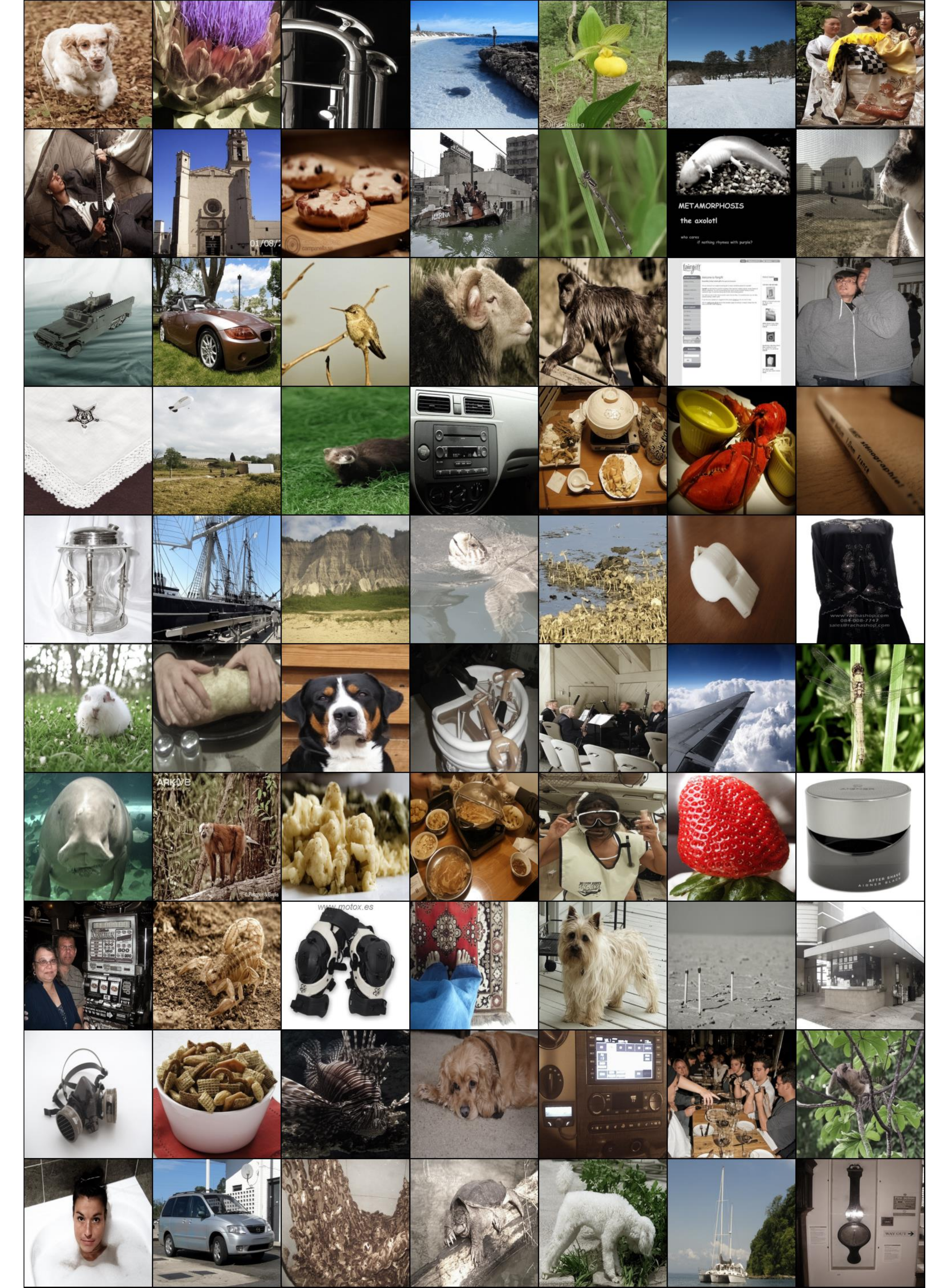}
    \caption{Uncurated images produced with a single hint where the hint location is randomly sampled from a uniform distribution. }
    \label{fig:uncurated_p1}
\end{figure*}

\begin{figure*}[t]
    \centering
    \includegraphics[width=0.92\textwidth]{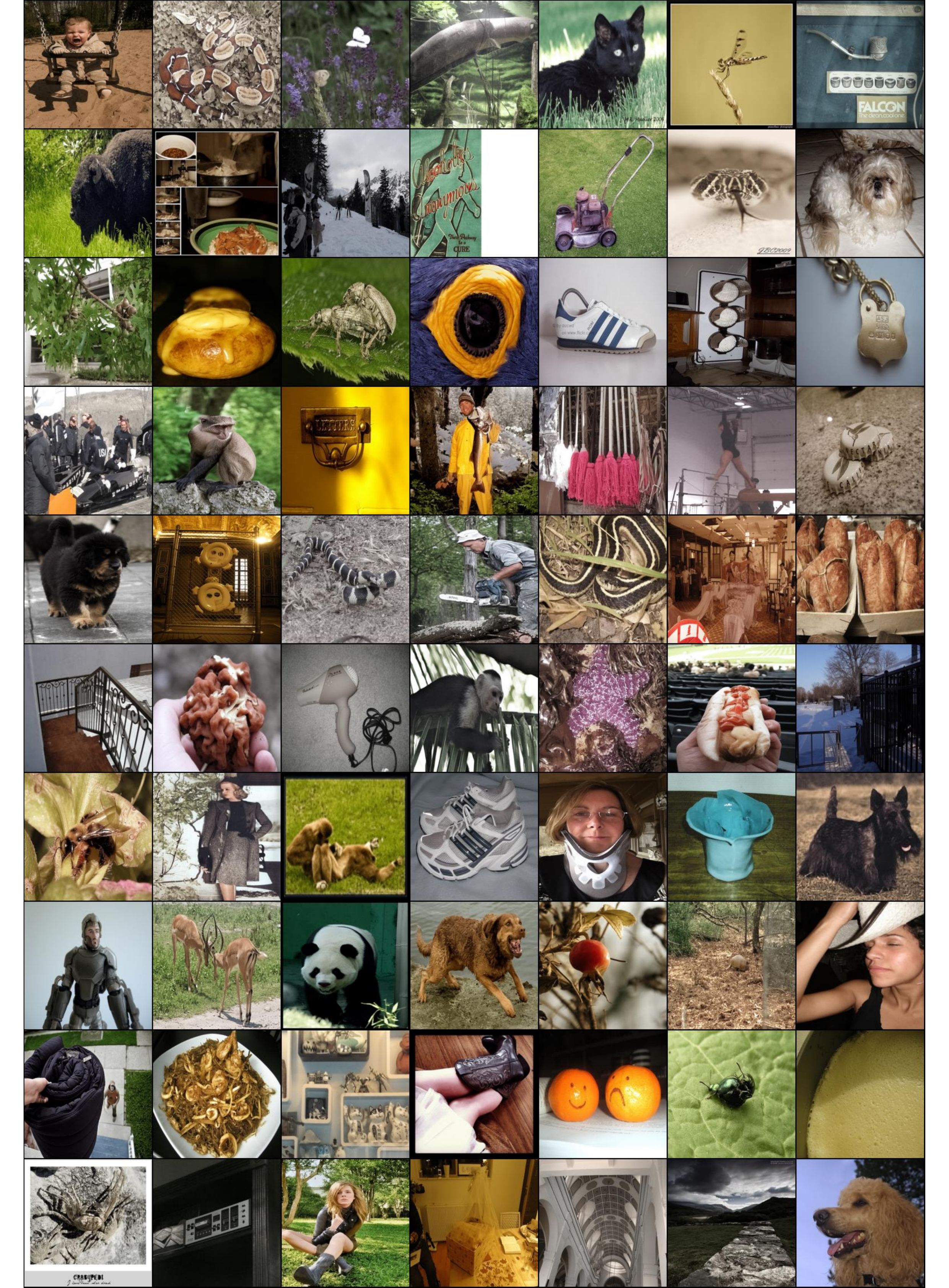}
    \caption{Uncurated images produced given five hints where the hint locations are randomly sampled from a uniform distribution.}
    \label{fig:uncurated_p5}
\end{figure*}

\begin{figure*}[t]
    \centering
    \includegraphics[width=0.92\textwidth]{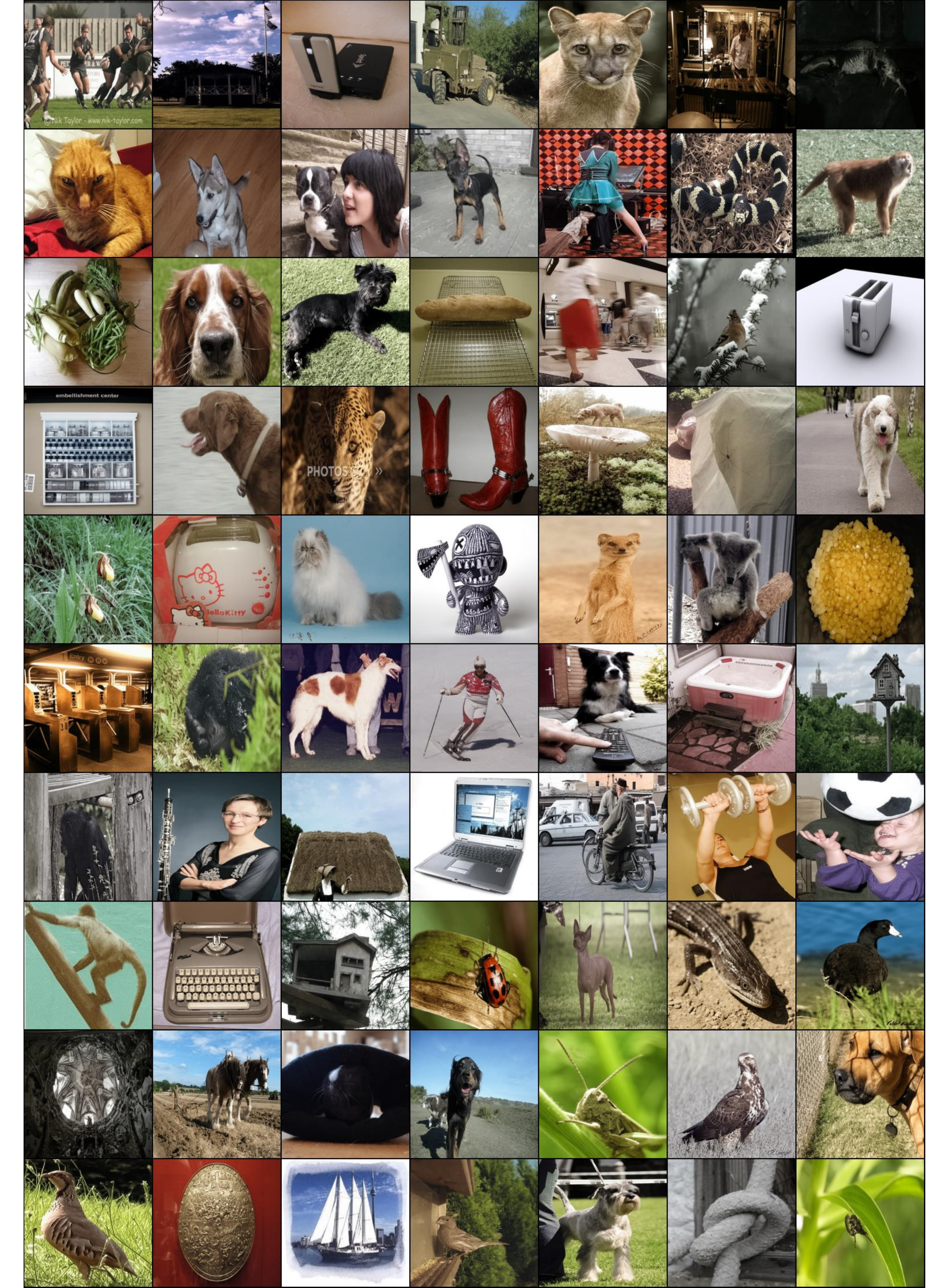}
    \caption{Uncurated images produced given ten hints where the hint locations are randomly sampled from a uniform distribution.}
    \label{fig:uncurated_p10}
\end{figure*}

\begin{figure*}[t]
    \centering
    \includegraphics[width=0.92\textwidth]{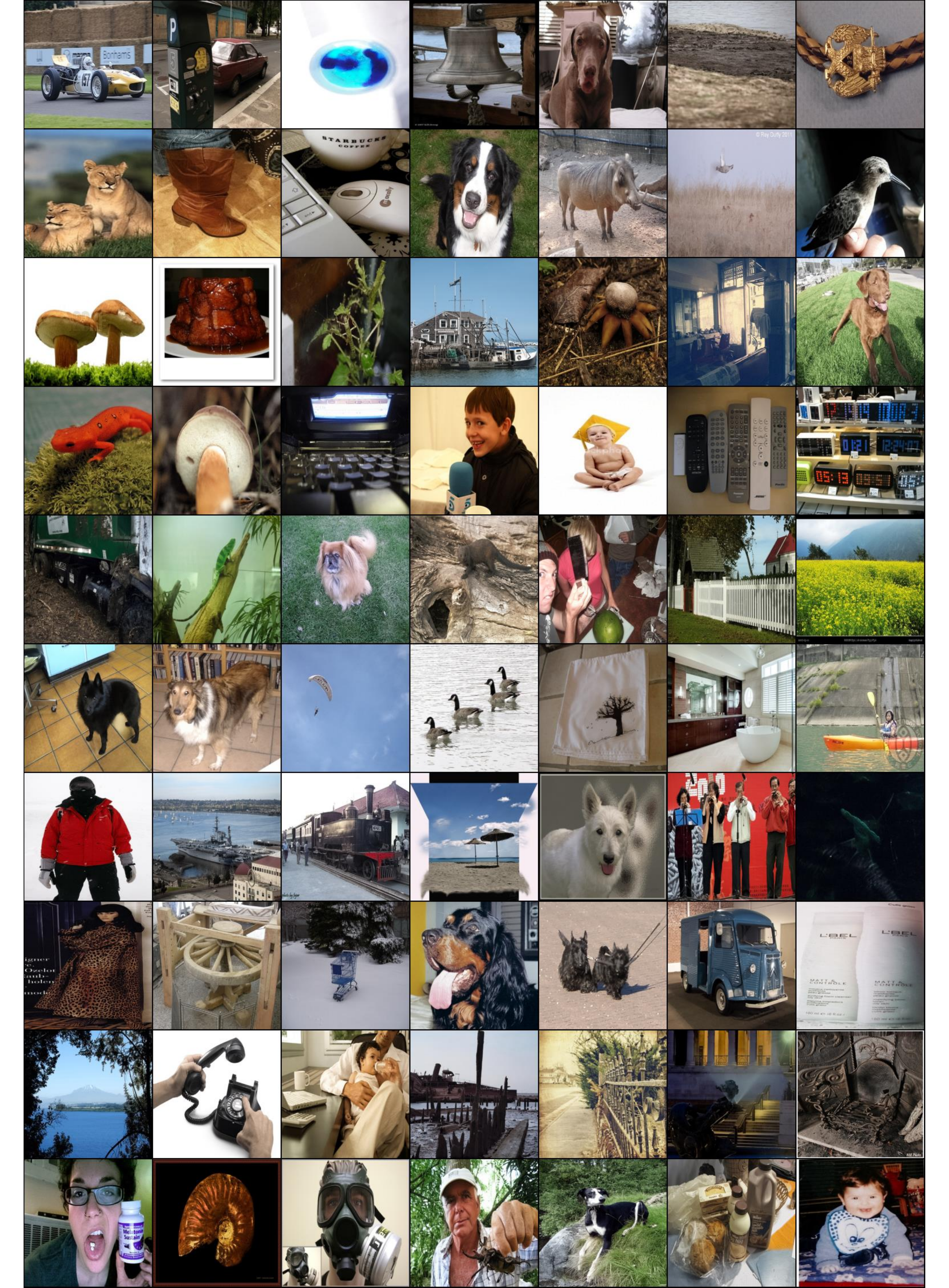}
    \caption{Uncurated images produced given hundred hints where the hint locations are randomly sampled from a uniform distribution.}
    \label{fig:uncurated_p100}
\end{figure*}

In this section and the remaining pages of the supplementary, we provide additional qualitative results. 
\cref{fig:add_quali_p1,fig:add_quali_p5,fig:add_quali_p10,fig:add_quali_p100} compares the results from different baseline models and iColoriT. 
Furthermore, uncurated results from the ImageNet ctest10k~\cite{ctest} of iColoriT are provided in \cref{fig:uncurated_p1,fig:uncurated_p5,fig:uncurated_p10,fig:uncurated_p100}. 
The figures are sorted according to the number of hints given to the model. 
We can observe that most images do not contain color-bleeding artifacts or partially colorized images unlike the baselines which produce partially colorized images. 
Finally, we attach a demo video demonstrating an interactive colorization scenario and a use case of iColoriT. 

\section{Full Details on the Quantitative Results}
\label{sec:sup_quanti_all}
In \Cref{tab:supp_quanti_psnr_ctest}, \Cref{tab:supp_quanti_psnr_flowers}, and \Cref{tab:supp_quanti_psnr_cub}, we provide the full details of the quantitative results for a fine-grained comparison. 



\newpage

\begin{table*}[h]
\begin{tabular}{@{}ccccccccc@{}}
\toprule
             & PSNR@1 & PSNR@2 & PSNR@5 & PSNR@10 & PSNR@20 & PSNR@50 & PSNR@100 & PSNR@200 \\ \midrule
Zhang~\etal~\cite{zhang2017} & 26.937 & 27.377 & 28.238 & 29.009  & 29.830  & 30.852  & 31.580   & 32.195   \\
Yin~\etal~\cite{side}   & 23.119 & 23.375 & 23.768 & 24.232  & 24.802  & 25.931  & 27.099   & 28.461   \\
Su~\etal~\cite{instanceaware}   & 27.275 & 27.656 & 28.422 & 29.108  & 29.833  & 30.734  & 31.370   & 31.886   \\ \midrule
iColoriT & \textbf{27.474} & \textbf{28.303} & \textbf{29.591} & \textbf{30.626} & \textbf{31.644} & \textbf{32.911} & \textbf{33.787} & \textbf{34.593} \\ \bottomrule
\end{tabular}
\caption{Full details of the PSNR achieved by each approach on the ImageNet ctest10k~\cite{ctest} dataset.}
\label{tab:supp_quanti_psnr_ctest}
\end{table*}

\begin{table*}[h]
\begin{tabular}{@{}ccccccccc@{}}
\toprule
      & PSNR@1          & PSNR@2 & PSNR@5 & PSNR@10 & PSNR@20 & PSNR@50 & PSNR@100 & PSNR@200 \\ \midrule
Zhang~\etal~\cite{zhang2017} & 22.720          & 23.270 & 24.250 & 25.130  & 25.930  & 27.010  & 27.826   & 28.665   \\
Yin~\etal~\cite{side}   & 18.452          & 18.617 & 18.893 & 19.445  & 19.937  & 21.075  & 22.362   & 24.082   \\
Su~\etal~\cite{instanceaware}    & \textbf{22.970} & 23.460 & 24.330 & 25.130  & 25.810  & 26.690  & 27.350   & 28.080   \\ \midrule
iColoriT & 22.925 & \textbf{24.190} & \textbf{26.044} & \textbf{27.370} & \textbf{28.384} & \textbf{29.742} & \textbf{30.731} & \textbf{31.756} \\ \bottomrule
\end{tabular}
\caption{Full details of the PSNR achieved by each approach on the Oxford 102flowers~\cite{flowers} dataset.}
\label{tab:supp_quanti_psnr_flowers}
\end{table*}

\begin{table*}[h]
\begin{tabular}{@{}ccccccccc@{}}
\toprule
      & PSNR@1 & PSNR@2 & PSNR@5 & PSNR@10 & PSNR@20 & PSNR@50 & PSNR@100 & PSNR@200 \\ \midrule
Zhang~\etal~\cite{zhang2017} & 27.450 & 27.900 & 28.660 & 29.320  & 29.980  & 30.880  & 31.570   & 32.180   \\
Yin~\etal~\cite{side}   & 23.547 & 23.936 & 24.661 & 25.097  & 25.647  & 26.621  & 27.623   & 28.876   \\
Su~\etal~\cite{instanceaware}    & 27.690 & 28.120 & 28.820 & 29.450  & 30.050  & 30.830  & 31.450   & 31.960   \\ \midrule
iColoriT & \textbf{27.986} & \textbf{28.782} & \textbf{29.806} & \textbf{30.595} & \textbf{31.462} & \textbf{32.634} & \textbf{33.543} & \textbf{34.453} \\ \bottomrule
\end{tabular}
\caption{Full details of the PSNR achieved by each approach on the CUB-200~\cite{cub} dataset.}
\label{tab:supp_quanti_psnr_cub}
\end{table*}

\newpage
\clearpage

\end{document}